\pdfoutput=1
\documentclass[10pt,twocolumn,letterpaper]{article}

\usepackage[pagenumbers]{cvpr}
\usepackage{booktabs}
\usepackage{graphicx}
\usepackage{placeins}
\usepackage{float}
\usepackage{makecell}
\usepackage{multirow}
\usepackage{array}
\usepackage{tabularx}
\usepackage{xcolor}
\usepackage{colortbl}

\newcolumntype{Z}[1]{>{\hsize=#1\hsize\linewidth=\hsize\centering\arraybackslash}X}

\renewcommand{\paragraph}[1]{\par\vspace{.5em}\noindent\textbf{#1}\enspace}

\definecolor{cvprblue}{rgb}{0.21,0.49,0.74}
\usepackage[pagebackref,breaklinks,colorlinks,allcolors=cvprblue]{hyperref}

\def\confName{CVPR}
\def\confYear{2027}

\title{Luniris: A Support-Conditioned Multimodal Foundation Model for Lunar Remote Sensing}

\makeatletter
\patchcmd{\@maketitle}{\vskip .375in}{\vskip .125in}{}{}
\patchcmd{\@maketitle}{\vspace*{24pt}}{\vspace*{12pt}}{}{}
\makeatother
\author{%
  Adam Ren\textsuperscript{1}\enspace Yuxuan Li\textsuperscript{1}\enspace Jinhan Li\textsuperscript{2}\enspace Huanzhang Qiu\textsuperscript{3}\enspace Zhisheng Chen\textsuperscript{1}\enspace Jinqi Zhao\textsuperscript{1}\enspace Zhong Lu\textsuperscript{1}\\[4pt]
  {\normalsize \textsuperscript{1}China University of Mining and Technology, Xuzhou}\\
  {\normalsize \textsuperscript{2}University of Science and Technology of China\qquad \textsuperscript{3}South China University of Technology}%
}
\hypersetup{
  pdftitle={Luniris: A Support-Conditioned Multimodal Foundation Model for Lunar Remote Sensing},
  pdfauthor={Adam Ren, Yuxuan Li, Jinhan Li, Huanzhang Qiu, Zhisheng Chen, Jinqi Zhao, Zhong Lu}
}

\begin{document}
\raggedbottom
\maketitle
\begin{abstract}

Multimodal lunar observations provide complementary information about surface morphology, composition, and physical properties, yet their coverage and effective spatial resolutions remain heterogeneous after spatial alignment. We propose Luniris, a lunar remote sensing foundation model that addresses both challenges through cross-modal masked pretraining. Evidence handles incomplete coverage by explicitly modeling hierarchical observation support to distinguish valid measurements from missing data and condition encoding, attention, and reconstruction target selection. Resolution-Supported Reconstruction (RSR) addresses resolution mismatch by adapting auxiliary supervision to the effective resolutions of targets and colocated valid visible observations, introducing coarser reconstruction constraints when the visible observations are coarser than the target. Evaluation on the Moonstone Benchmark and SomBench spans ten tasks covering classification, regression, segmentation, and detection. Luniris achieves the best performance among the evaluated models on multiple tasks and remains competitive across both benchmarks. Matched-data comparisons and component ablations further support the benefits of its pretraining design.

\end{abstract}
\section{Introduction}
\label{sec:introduction}

The Moon, Earth's only natural satellite, is a major target for extraterrestrial resource exploration~\cite{crawford2015resources}. Over its history, solar wind irradiation, meteoroid impacts, volcanism, and tectonic deformation have shaped the morphology, physical properties, and composition of its surface, leaving a rich record of lunar evolution~\cite{jaumann2012geology,pieters2016space}. From early telescopic observations in 1609~\cite{chapman2009harriot} to modern orbital remote sensing, lunar observations have become a key means of understanding the Moon's thermal history and the external processes that modify its surface~\cite{hiesinger2006new}. Successive orbital missions have supported resource assessment, exploration planning, and research on surface evolution by providing complementary measurements of morphology and topography~\cite{robinson2010lroc,smith2010lola}, spectral reflectance and thermophysical properties~\cite{pieters2009m3,paige2010diviner}, and radar scattering and gravity~\cite{nozette2010minirf,zuber2013gravity}. Learning transferable representations from these multimodal lunar observations provides a reusable foundation for studying surface evolution, assessing resource distributions, and selecting exploration sites.

\begin{figure}[t]
    \centering
    \includegraphics[width=\columnwidth]{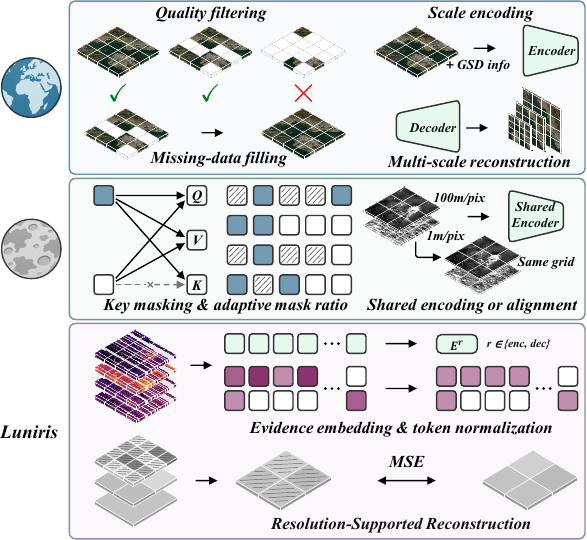}
    \caption{Strategies for handling missing observations and resolution differences in remote sensing foundation models. Luniris conditions representations on observation support and selects auxiliary reconstruction scales using the effective resolutions of the target and valid visible observations within the same spatial patch.}
    \label{fig:overview}
\end{figure}

Self-supervised visual pretraining enables transferable representations to be learned from unlabeled observations~\cite{caron2021emerging,he2022masked,oquab2024dinov2}, while multimodal masked modeling exploits relationships between complementary inputs and reconstruction targets~\cite{bachmann2022multimae,mizrahi2023massively}.
Earth observation models extend these ideas to temporal and spectral imagery~\cite{cong2022satmae,szwarcman2026prithvi,hong2024spectralgpt}, cross-sensor representation learning~\cite{fuller2023croma,xiong2025dofa,guo2024skysense}, and multimodal prediction and generation~\cite{nedungadi2024mmearth,jakubik2025terramind}.
Recent lunar foundation models demonstrate the value of learning directly from lunar observations.
Moonstone introduces modality-grouped pretraining with coverage handling~\cite{prasad2026moonstone}; LunarFM learns shared representations for scientific analysis~\cite{girona2026lunarfm}; and the NASA--IBM Lunar Foundation Model combines acquisition geometry with mixed-resolution pretraining~\cite{fraccaro2026lfm}.
These advances establish a basis for learning across lunar instruments and tasks.

A central challenge remains within spatially aligned observations: \emph{a shared grid does not imply uniform observation support or equal spatial resolving power}. Figure~\ref{fig:overview} contrasts representative strategies for addressing these issues and highlights the two core designs of Luniris.
Channel-dependent coverage produces fully observed, partially supported, and missing regions within the same input crop.
Their support patterns persist after alignment and determine which measurements can contribute to encoding and reconstruction.
Meanwhile, resampling places source products on a common grid without equalizing their effective resolutions.
Cross-modal masking makes this distinction especially consequential: a masked fine-resolution target may be accompanied only by coarse-resolution unmasked observations at the same spatial patch.
Scale-aware encoding and multiscale reconstruction provide useful tools for heterogeneous imagery~\cite{reed2023scale,noman2024rethinking,astruc2025anysat}.
We bring observation support and effective resolution into a common pretraining formulation, explicitly conditioning representations on available support and selecting auxiliary reconstruction scales from the observations that remain visible.

We introduce \textbf{Luniris}, a lunar foundation model pretrained through cross-modal masked reconstruction.
Building on modality-grouped tokenization~\cite{prasad2026moonstone}, we organize inputs into physical modality--spatial lattices, each containing all channels of one modality within a spatial patch, and perform tokenization and masking on these lattices.
\textbf{Evidence} summarizes pixelwise observation support at channel, lattice, and crop levels and supplies hierarchical support embeddings to the encoder and decoder.
Observation support also normalizes token content, determines attention-key eligibility, and restricts reconstruction supervision to supported target pixels.
\textbf{Resolution-Supported Reconstruction (RSR)} selects an auxiliary supervision scale from the effective resolutions of each target channel and valid unmasked observations within the same spatial patch.
When the finest available observations are coarser than the target, RSR compares spatially averaged predictions and targets at the coarser scale.
This auxiliary objective complements modality-balanced pixel-level reconstruction, with spectral and Stokes physical regularization completing the pretraining objective.

We pretrain Luniris on a curated 30-channel corpus spanning seven physical modalities, replacing the original Moonstone radar inputs with four Miniature Radio Frequency (Mini-RF) Stokes channels.
Evaluation on the Moonstone Benchmark~\cite{prasad2026moonstone} and SomBench~\cite{patil2026sombench} covers ten classification, regression, segmentation, and detection tasks.
Luniris achieves competitive performance across both benchmarks, with task-specific advantages over the evaluated Earth and lunar foundation models.
Comparisons with a vanilla masked autoencoder (MAE) pretrained on the same corpus support the benefits of the pretraining design, while ablations examine the contributions of hierarchical support embeddings and RSR.
Our contributions are threefold:
\begin{itemize}
    \item We introduce Luniris, which integrates hierarchical observation support into cross-modal lunar representation learning through support-conditioned encoding and reconstruction.
    \item We propose RSR, an auxiliary reconstruction objective whose supervision scale depends on target effective resolution and valid unmasked observations within the same spatial patch.
    \item We evaluate transfer across two lunar benchmarks and ten tasks using Earth and lunar baselines, matched-data controls, and component ablations.
\end{itemize}
\section{Related Work}
\label{sec:related_work}

\subsection{Earth Remote Sensing Foundation Models}

Earth observation foundation models have expanded from spectral and temporal pretraining to learning across heterogeneous sensors and modalities. SatMAE~\cite{cong2022satmae} incorporates temporal encoding and spectral grouping into masked modeling, while SpectralGPT~\cite{hong2024spectralgpt} learns joint spatial--spectral dependencies. CROMA~\cite{fuller2023croma} and SkySense~\cite{guo2024skysense} combine cross-modal learning and fusion of radar and optical observations, and DOFA~\cite{xiong2025dofa} uses wavelength-conditioned weights for different spectral configurations. MMEarth~\cite{nedungadi2024mmearth} broadens pretraining supervision through heterogeneous geospatial targets, while TerraMind~\cite{jakubik2025terramind} integrates multimodal representation learning and generation.

Incomplete inputs are handled through data curation, imputation, and flexible modality use. Prithvi-EO-2.0~\cite{szwarcman2026prithvi} filters samples with excessive missingness or cloud cover and interpolates remaining missing pixels. MultiMAE~\cite{bachmann2022multimae} establishes cross-modal masked prediction with flexible input subsets, and TerraMind~\cite{jakubik2025terramind} generates additional modalities from available observations. AnySat~\cite{astruc2025anysat} learns sensor-specific scalar values to fill missing channels during downstream adaptation. These mechanisms address distinct aspects of incomplete observations, from missing measurements to unavailable modalities.

Resolution heterogeneity has motivated scale-aware representations and reconstruction objectives. Scale-MAE~\cite{reed2023scale} combines ground-sampling-distance-aware positional encoding with low- and high-frequency reconstruction, and SatMAE++~\cite{noman2024rethinking} introduces multiscale reconstruction. AnySat~\cite{astruc2025anysat} represents patches covering the same physical area through scale-adaptive encoders that accommodate different pixel dimensions. OlmoEarth~\cite{herzog2026olmoearth} resamples inputs to a common grid while retaining native-resolution band groups for tokenization and cross-bandset prediction. Luniris adapts auxiliary supervision scales to the effective resolutions of the target and colocated observations that remain valid and visible after masking.

\subsection{Lunar Remote Sensing Foundation Models}

Lunar foundation models learn shared representations from complementary orbital products. Moonstone~\cite{prasad2026moonstone} establishes a multimodal corpus and benchmark with a modality-grouped masked autoencoder (MG-MAE), which combines physical-modality tokenization and shared encoding with missing-token attention masking, coverage-adaptive masking ratios, and reconstruction weighting by valid-pixel fraction. LunarFM~\cite{girona2026lunarfm} adopts a MultiMAE-style architecture to construct lunar embeddings for scientific analysis, filling missing inputs with zeros and excluding them from reconstruction loss. The NASA--IBM Lunar Foundation Model~\cite{fraccaro2026lfm} incorporates acquisition geometry into multimodal pretraining and jointly trains on meter- and hundred-meter-scale tiles, sampling input and target tokens only from modalities present in each sample. Luniris extends coverage handling to explicit support-conditioned representation learning across aligned lunar modalities.
\section{Method}
\label{sec:method}

\begin{figure*}[t]
    \centering
    \includegraphics[width=\textwidth]{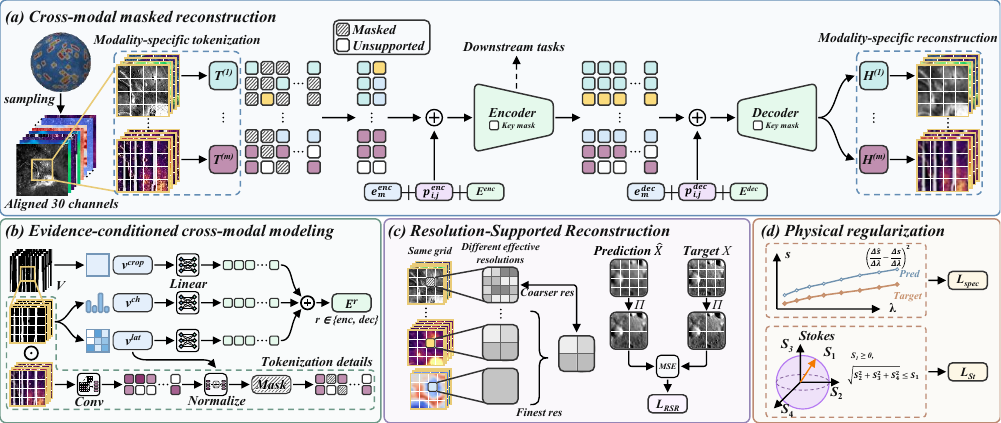}
    \caption{Overview of Luniris. (a) Aligned lunar observations are tokenized and masked as modality--spatial lattices for cross-modal reconstruction, and the pretrained encoder transfers to downstream tasks. (b) Evidence uses observation support to weight token content, construct hierarchical support embeddings, and determine attention-key validity. (c) RSR selects auxiliary supervision scales from the target and valid unmasked observations within the same spatial patch, then compares prediction and target means over supported pixels. (d) Physical regularization matches adjacent-band Moon Mineralogy Mapper (M$^3$) reflectance slopes and constrains radar Stokes predictions.}
    \label{fig:framework}
\end{figure*}

Luniris is a lunar foundation model pretrained through cross-modal masked reconstruction (\cref{fig:framework}). Tokenization and masking use physical modality--spatial lattices. Evidence conditions encoding and reconstruction target selection on observation support, while Resolution-Supported Reconstruction (RSR) determines auxiliary supervision scales from the effective resolutions of targets and unmasked observations within the same patch.

\subsection{Problem Formulation}
\label{sec:problem_formulation}

Luniris pretrains an encoder on multimodal lunar observations for transfer to downstream tasks. Given aligned observations $\{X_m\}_{m=1}^{G}$ from $G$ physical modalities, with $X_m\in\mathbb{R}^{C_m\times H\times W}$, each modality--spatial lattice $X_{m,i,j}\in\mathbb{R}^{C_m\times P\times P}$ contains all $C_m$ channels of modality $m$ within spatial patch $(i,j)$. Downstream prediction is formulated as
\begin{equation}
\widehat{Y}_{\tau}=f_{\tau}\!\left(\Phi_{\theta}^{\mathrm{Luniris}}\!\left(\{X_{m,i,j}\}_{m,i,j},V\right)\right),
\label{eq:luniris_transfer}
\end{equation}
where $\Phi_{\theta}^{\mathrm{Luniris}}$ is the pretrained encoder, $f_{\tau}$ is the predictor for downstream task $\tau$, and $\widehat{Y}_{\tau}$ is its prediction. $V$ records pixelwise observation support for each channel on the aligned grid.

\subsection{Evidence-Conditioned Cross-Modal Modeling}
\label{sec:Evidence-Conditioned Cross-Modal Modeling}

Lunar observations are organized into modality--spatial lattices, with explicit observation support conditioning encoding and reconstruction.

\paragraph{Physical-modality tokenization and masking.}
For the same lunar surface region, different remote sensing modalities provide distinct yet complementary physical information. For example, spectral reflectance informs lunar surface composition, while radar Stokes measurements are sensitive to dielectric properties and roughness. The lattices form a $G\times H_c\times W_c$ array, where $H_c=H/P$ and $W_c=W/P$, and are mapped to tokens through modality-specific tokenization. Lattices are randomly masked at ratio $\rho$, yielding unmasked and masked sets $\mathcal{O}$ and $\mathcal{M}$. Lattices from different modalities at the same spatial patch can be masked separately, encouraging reconstruction from complementary colocated modalities and observations at other spatial locations.

\paragraph{Hierarchical observation support.}
Spatially aligned multisource lunar remote sensing observations still exhibit channel-dependent patterns of missing data. For channel $c$ at pixel $(u,v)$ within lattice $X_{m,i,j}$, observation support is represented by $V_{m,i,j,c,u,v}\in[0,1]$, where zero indicates missing data and fractional values indicate partial support. Evidence summarizes this support at channel, lattice, and crop levels:
\begin{equation}
\begin{aligned}
v_{m,i,j,c}^{\mathrm{ch}} &= \frac{1}{P^2}\sum_{u,v}V_{m,i,j,c,u,v},\\
v_{m,i,j}^{\mathrm{lat}} &= \frac{1}{C_m}\sum_{c=1}^{C_m}v_{m,i,j,c}^{\mathrm{ch}},\\
v_m^{\mathrm{crop}} &= \frac{1}{H_cW_c}\sum_{i,j}v_{m,i,j}^{\mathrm{lat}}.
\end{aligned}
\label{eq:evidence_support}
\end{equation}
Here $v_{m,i,j,c}^{\mathrm{ch}}$ represents the average pixelwise observation support for channel $c$ within a lattice. The lattice-level summary $v_{m,i,j}^{\mathrm{lat}}$ averages support across the $C_m$ channels of that lattice, while the crop-level summary $v_m^{\mathrm{crop}}$ averages support for modality $m$ across all $H_cW_c$ spatial patches in the input crop. The three summaries are computed before artificial masking and retained for both masked and unmasked lattices.

\paragraph{Support-conditioned encoding and decoding.}
Partially observed lattices contain both measurements and missing-data placeholders. Observation support weights token content and determines attention-key eligibility. Let $\boldsymbol{v}_{m,i,j}^{\mathrm{ch}}=[v_{m,i,j,c}^{\mathrm{ch}}]_{c=1}^{C_m}$ collect the channel-level support values. Unsupported input entries are set to zero. Here $h_{m,i,j}$ is the support-normalized content representation of lattice $X_{m,i,j}$, and $\operatorname{Conv}_m$ is a modality-specific convolutional tokenizer with kernel size and stride $P$. The index $r\in\{\mathrm{enc},\mathrm{dec}\}$ distinguishes encoder and decoder branches. The constant $\epsilon$ stabilizes support normalization. We parameterize support embeddings using the support summaries and construct encoder input tokens as
\begin{equation}
\begin{aligned}
h_{m,i,j} &= \frac{\mathbf{1}[v_{m,i,j}^{\mathrm{lat}}>0]\,\operatorname{Conv}_m(X_{m,i,j}\odot V_{m,i,j})}{\max(v_{m,i,j}^{\mathrm{lat}},\epsilon)},\\
E_{m,i,j}^{r}(V) &= \mathcal{P}_{m,r}^{\mathrm{ch}}(\boldsymbol{v}_{m,i,j}^{\mathrm{ch}})+\mathcal{P}_r^{\mathrm{lat}}(v_{m,i,j}^{\mathrm{lat}})+\mathcal{P}_r^{\mathrm{crop}}(v_m^{\mathrm{crop}}),\\
t_{m,i,j} &= h_{m,i,j}+e_m^{\mathrm{enc}}+p_{i,j}^{\mathrm{enc}}+E_{m,i,j}^{\mathrm{enc}}(V).
\end{aligned}
\label{eq:evidence_tokens}
\end{equation}
The encoder and decoder branches use separate learned linear projections $\mathcal{P}$. The encoder input token $t_{m,i,j}$ combines the support-normalized content representation $h_{m,i,j}$ with three embeddings: the modality embedding $e_m^{\mathrm{enc}}$, the position embedding $p_{i,j}^{\mathrm{enc}}$, and the hierarchical observation support embedding $E_{m,i,j}^{\mathrm{enc}}(V)$.

The encoder $\Phi_{\theta}^{\mathrm{Luniris}}$ combines the token construction above with a base-sized Vision Transformer (ViT-B) backbone. During pretraining, it encodes the unmasked tokens as
\begin{equation}
Z=\operatorname{Enc}_{\theta}\!\left(\{t_{m,i,j}\}_{(m,i,j)\in\mathcal{O}};K_{\mathcal{O}}(V)\right).
\label{eq:luniris_encoding}
\end{equation}
Here, $K_{\mathcal{O}}(V)$ determines attention-key validity for the unmasked lattices. A lattice is masked as an attention key when none of its pixels is jointly valid across all channels. The decoder adds hierarchical support embeddings and uses the same validity mask for these unmasked lattices, while retaining mask tokens as attention keys. Modality-specific reconstruction heads predict the masked lattices, with supervision restricted to pixels where the target observations are supported.

\subsection{Resolution-Supported Reconstruction}
\label{sec:rscr}

Lunar remote sensing modalities vary widely in spatial resolution and undergo upsampling or downsampling during spatial alignment, leaving different effective resolutions on the shared grid. Cross-modal masking can therefore require reconstructing masked fine-resolution modalities from visible coarse-resolution modalities. In this case, RSR adds an auxiliary loss that compares the low-frequency content of predictions and targets through spatial averaging, with the pooling grid determined by the effective resolution of the visible coarse observations.

\paragraph{Supervision scale selection.}
For a masked lattice at spatial patch $(i,j)$, RSR first finds the finest resolution among valid channels in unmasked lattices at the same patch, denoted by $g_{i,j}^{\mathrm{vis}}$. It then compares this value with the resolution $g_{m,c}$ of each target channel and uses the coarser one, $g_{m,c,i,j}^{\mathrm{sup}}$, for auxiliary supervision. $g_{m,c}$ is the larger of the channel's registered source resolution and the shared grid spacing $\delta$. Let $\mathcal{C}_{i,j}$ collect channels with valid pixels from unmasked lattices in patch $(i,j)$, indexed by modality $n$ and channel $c'$. When $\mathcal{C}_{i,j}$ is nonempty, the finest visible resolution and the auxiliary supervision scale are
\begin{equation}
\begin{aligned}
g_{i,j}^{\mathrm{vis}} &= \min_{(n,c')\in\mathcal{C}_{i,j}}g_{n,c'},\\
g_{m,c,i,j}^{\mathrm{sup}} &= \max\!\left(g_{m,c},g_{i,j}^{\mathrm{vis}}\right).
\end{aligned}
\label{eq:rscr_scale}
\end{equation}
When the finest valid unmasked observations within the same patch are at least as fine as the target channel, RSR sets the auxiliary supervision scale to the effective resolution of that target channel. When those observations are coarser than the target, RSR instead uses the finest effective resolution among the valid unmasked observations in the patch. The selected resolution determines the pooling grid. Prediction and target means are then compared over the regions where this grid intersects the target spatial patch.

\paragraph{Lattice-local projection and reconstruction.}
Predictions and targets are spatially averaged before computing the RSR loss. The pooling factor $k_{m,c,i,j}$ is obtained by rounding $g^{\mathrm{sup}}_{m,c,i,j}/\delta$ to the nearest integer and clipping it to $[1,\min(H,W)]$. Adaptive average pooling $\mathcal{A}_k$ then maps the full image to $\lceil H/k\rceil\times\lceil W/k\rceil$ regions.

Within each pooling region $q$, predictions and targets are averaged over the same pixels inside the target patch $(i,j)$ with positive target-channel support ($V>0$). Let $\Omega_q$ denote these pixels, $p$ index pixel locations, and $U$ denote the prediction or target channel map. The mean is
\begin{equation}
\Pi_{m,c,i,j}(U;V)_q=\frac{1}{|\Omega_q|}\sum_{p\in\Omega_q}U(p).
\label{eq:rscr_projection}
\end{equation}
Regions with no observed target pixels are skipped. The selected resolution sets the pooling grid, while each average is restricted to the target patch; its spatial support is therefore bounded by the patch extent.

For target $s$, let $\Pi_s$ denote the averaging operator in \cref{eq:rscr_projection}, and let $\hat{\mu}_{s,q}$ and $\mu_{s,q}$ denote its prediction and target means in pooling region $q$.
\begin{equation}
\mathcal{L}_{\mathrm{RSR}}=\frac{1}{|\mathcal{S}|}\sum_{s\in\mathcal{S}}\frac{1}{|\mathcal{Q}_s|}\sum_{q\in\mathcal{Q}_s}(\hat{\mu}_{s,q}-\mu_{s,q})^2.
\label{eq:rscr_loss}
\end{equation}
Here each target $s$ is a channel in a masked lattice from an input crop. $\mathcal{Q}_s$ contains pooling regions with valid target pixels. $\mathcal{S}$ includes only targets with valid pixels whose spatial patch $(i,j)$ has a nonempty visible-channel set $\mathcal{C}_{i,j}$.

\subsection{Pretraining Objective}
\label{sec:pretraining_objective}

Pretraining combines modality-balanced reconstruction, the RSR loss defined in \cref{eq:rscr_loss}, and physical regularization.

\paragraph{Modality-balanced reconstruction.}
The reconstruction loss $\mathcal{L}_{\mathrm{rec}}$ is computed within each local microbatch in three averaging steps. First, normalized-space squared errors are averaged over valid masked pixels separately for each channel. The channel losses are then averaged with equal weights within each modality, followed by an equal-weight average across modalities. Channels without valid targets are excluded, as are modalities with no valid targets.

\paragraph{Physical regularization.}
After inverse normalization, $\mathcal{L}_{\mathrm{spec}}$ matches adjacent-band M$^3$ reflectance slopes, while $\mathcal{L}_{\mathrm{St}}$ penalizes negative radar intensity and physically infeasible polarization. The total objective is
\begin{equation}
\begin{aligned}
\mathcal{L}_{\mathrm{total}} &= \mathcal{L}_{\mathrm{rec}}+\lambda_{\mathrm{RSR}}\mathcal{L}_{\mathrm{RSR}}\\
&\quad+\lambda_{\mathrm{phys}}\alpha(t)(\mathcal{L}_{\mathrm{spec}}+\mathcal{L}_{\mathrm{St}}),
\end{aligned}
\label{eq:pretraining_objective}
\end{equation}
where the $\lambda$ coefficients weight auxiliary losses, $t$ counts loss evaluations, and $\alpha(t)$ gradually increases the physical regularization weight. Loss terms with no valid targets are set to zero. Detailed loss definitions, validity conditions, and the weighting schedule are provided in Appendix~A. The encoder transfers to downstream tasks via \cref{eq:luniris_transfer}.
\section{Experiments}
\label{sec:experiments}

\subsection{Experimental Setup}
\label{sec:experimental_setup}

\paragraph{Datasets and Data Splits.} We pretrain Luniris on Moonstone~\cite{prasad2026moonstone}, replacing its two radar channels with four Mini-RF S-band zoom-mode Stokes channels that we collected, preprocessed, and curated. The resulting corpus contains 30 channels across seven physical modality groups, with valid Stokes observations covering approximately 67\% of pixels on the global lunar grid.

Spherical Sampling and Partitioning (SSP) divides the lunar surface into 768 equal-area Hierarchical Equal Area isoLatitude Pixelation (HEALPix)~\cite{gorski2005healpix} regions at \(N_{\mathrm{side}}=8\). Of these, 614 regions are assigned to training, 77 to validation, and 77 to testing. During pretraining, crops are sampled online within the training regions, using a sampling density that is uniform per unit lunar surface area. Channel specifications, preprocessing, and sampling details are provided in the supplementary material.

Downstream evaluation uses the Moonstone Benchmark and SomBench~\cite{patil2026sombench}. For Moonstone, we retain the original task definitions, spatially realign the inputs, and use identical SSP splits for all models. SomBench follows the official datasets and splits.

\paragraph{Implementation Details.} Luniris uses a ViT-B encoder with \(256\times256\) inputs, \(16\times16\) patches, and 75\% masking. Pretraining runs for 150 epochs of 128,000 crops with an effective batch size of 64, using AdamW, a peak learning rate of \(10^{-4}\), linear warmup, and cosine decay. All training runs are conducted on four NVIDIA H100 GPUs.

All downstream models use a unified experimental framework, with hyperparameters and checkpoints selected using validation-set metrics. We reproduce Moonstone MG-MAE~\cite{prasad2026moonstone} from its public data and code, adding mechanisms described in the paper but absent from the released code.

\subsection{Pretraining Reconstruction}
\label{sec:Pretraining Reconstruction}

\begin{figure}[t]
    \centering
    \includegraphics[width=\linewidth]{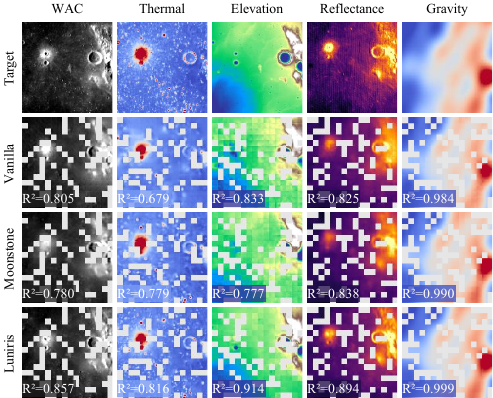}
    \caption{Selected-channel reconstructions. Columns: LROC WAC morphology, Diviner midnight bolometric temperature, Lunar Orbiter Laser Altimeter blended elevation, M$^3$ 750\,nm reflectance, and Gravity Recovery and Interior Laboratory (GRAIL) free-air gravity anomaly. Rows: target, Vanilla MAE, Moonstone MG-MAE, and Luniris. Insets report $R^2$.}
    \label{fig:reconstruction}
\end{figure}

Figure~\ref{fig:reconstruction} compares reconstructions of selected Lunar Reconnaissance Orbiter Camera (LROC) Wide Angle Camera (WAC) morphology, thermal, elevation, M$^3$ 750\,nm reflectance, and GRAIL free-air gravity anomaly channels. Across these examples, Luniris reconstructs spatial structures and intensity variations more faithfully than Vanilla MAE and Moonstone MG-MAE. In WAC, it better distinguishes dark crater interiors from bright rims, while the thermal and reflectance reconstructions preserve more distinct localized high-intensity regions. Its elevation reconstruction more closely follows the broad terrain gradient while keeping small crater depressions spatially localized. For GRAIL free-air gravity anomaly, it better preserves the high-value region and its gradual transition to the surrounding lower-value field. Panel scores report the standard coefficient of determination ($R^2$) over valid pixels jointly supported by the reconstruction mask and observation evidence. Reconstructions for the remaining channels are provided in the supplementary material.

\subsection{Comparison with Existing Methods}
\label{sec:comparison}

\begin{table*}[t]
\centering
\caption{Downstream comparison on Moonstone Benchmark and SomBench. Adaptation settings are indicated in task labels. Bold/underlined values mark the best/second-best results within each setting, ranked before rounding. Complete results, model versions, and adaptation details are provided in the supplementary material.}
\label{tab:benchmark_combined}

\begingroup
\small
\setlength{\tabcolsep}{1pt}
\setlength{\aboverulesep}{0.4pt}
\setlength{\belowrulesep}{0.4pt}
\renewcommand{\arraystretch}{0.90}

\begin{tabular}{@{}ll*{10}{c}@{}}
\toprule
& & Control & \multicolumn{6}{c}{Earth FMs} & \multicolumn{2}{c}{Lunar FMs} & Ours \\
\cmidrule(l{3pt}r{3pt}){3-3}
\cmidrule(l{3pt}r{3pt}){4-9}
\cmidrule(l{3pt}r{3pt}){10-11}
\cmidrule(l{3pt}r{1pt}){12-12}
Task & Metric & Vanilla~MAE & SatMAE & DOFA & TerraMind & Prithvi & SpectralGPT+ & CROMA & Moonstone & NASA--IBM & \textbf{Luniris} \\
\midrule

\rowcolor{black!5}\multicolumn{12}{l}{\textbf{Moonstone Benchmark}} \\
\midrule
\multirow{2}{*}{Geo (LP)} & \mbox{Acc $\uparrow$} & \underline{0.335} & 0.306 & 0.304 & 0.282 & 0.291 & 0.284 & 0.255 & 0.327 & 0.285 & \textbf{0.379} \\
& \mbox{mF1 $\uparrow$} & \underline{0.193} & 0.183 & 0.126 & 0.157 & 0.139 & 0.124 & 0.109 & 0.163 & 0.132 & \textbf{0.216} \\
\arrayrulecolor{black!25}\cmidrule[0.25pt]{1-12}\arrayrulecolor{black}

\multirow{2}{*}{Age (FFT)} & \mbox{Acc $\uparrow$} & \underline{0.517} & 0.501 & 0.515 & 0.496 & 0.510 & 0.483 & 0.481 & 0.505 & 0.484 & \textbf{0.532} \\
& \mbox{mF1 $\uparrow$} & 0.368 & 0.362 & 0.355 & 0.321 & \underline{0.398} & 0.328 & 0.381 & 0.391 & \textbf{0.413} & 0.377 \\
\arrayrulecolor{black!25}\cmidrule[0.25pt]{1-12}\arrayrulecolor{black}

\multirow{2}{*}{Comp (LP)} & \mbox{m$R^2$ $\uparrow$} & \textbf{0.922} & 0.875 & 0.824 & 0.886 & 0.884 & 0.889 & 0.878 & 0.884 & 0.891 & \underline{0.913} \\
& \mbox{mRMSE $\downarrow$} & \textbf{0.277} & 0.353 & 0.417 & 0.336 & 0.340 & 0.332 & 0.347 & 0.339 & 0.329 & \underline{0.294} \\
\arrayrulecolor{black!25}\cmidrule[0.25pt]{1-12}\arrayrulecolor{black}

\multirow{2}{*}{Therm (LP)} & \mbox{m$R^2$ $\uparrow$} & 0.907 & 0.891 & 0.757 & 0.891 & 0.903 & 0.859 & \underline{0.912} & 0.903 & 0.833 & \textbf{0.923} \\
& \mbox{mRMSE $\downarrow$} & 0.326 & 0.354 & 0.527 & 0.351 & 0.331 & 0.401 & \underline{0.317} & 0.331 & 0.436 & \textbf{0.302} \\
\arrayrulecolor{black!25}\cmidrule[0.25pt]{1-12}\arrayrulecolor{black}

\multirow{2}{*}{Mare (LP)} & \mbox{mIoU $\uparrow$} & \underline{0.919} & 0.909 & 0.891 & 0.910 & 0.901 & 0.825 & 0.913 & 0.916 & 0.906 & \textbf{0.920} \\
& \mbox{mF1 $\uparrow$} & \underline{0.957} & 0.951 & 0.940 & 0.952 & 0.946 & 0.898 & 0.953 & 0.955 & 0.949 & \textbf{0.957} \\
\arrayrulecolor{black!25}\cmidrule[0.25pt]{1-12}\arrayrulecolor{black}

\multirow{2}{*}{Crater (FFT)} & \mbox{mIoU $\uparrow$} & 0.656 & 0.614 & 0.572 & 0.603 & 0.610 & 0.604 & 0.617 & 0.631 & \underline{0.666} & \textbf{0.693} \\
& \mbox{mF1 $\uparrow$} & 0.789 & 0.757 & 0.721 & 0.748 & 0.754 & 0.748 & 0.759 & 0.771 & \underline{0.795} & \textbf{0.816} \\
\specialrule{0.6pt}{1pt}{0pt}

\rowcolor{black!5}\multicolumn{12}{l}{\textbf{SomBench}} \\
\midrule
\multirow{2}{*}{WAC (LoRA)} & \mbox{mAP $\uparrow$} & 0.168 & 0.170 & 0.174 & 0.172 & 0.175 & 0.156 & 0.171 & 0.177 & \textbf{0.257} & \underline{0.241} \\
& \mbox{mAP50 $\uparrow$} & 0.325 & 0.340 & 0.351 & 0.336 & 0.336 & 0.309 & 0.335 & 0.351 & \textbf{0.613} & \underline{0.576} \\
\arrayrulecolor{black!25}\cmidrule[0.25pt]{1-12}\arrayrulecolor{black}

\multirow{2}{*}{NAC (FFT)} & \mbox{mAP $\uparrow$} & 0.081 & 0.085 & 0.101 & 0.079 & 0.090 & 0.074 & 0.085 & 0.093 & \underline{0.164} & \textbf{0.165} \\
& \mbox{mAP50 $\uparrow$} & 0.206 & 0.209 & 0.248 & 0.199 & 0.219 & 0.189 & 0.216 & 0.228 & \underline{0.500} & \textbf{0.502} \\
\arrayrulecolor{black!25}\cmidrule[0.25pt]{1-12}\arrayrulecolor{black}

\multirow{2}{*}{IMP (LoRA)} & \mbox{mIoU $\uparrow$} & 0.553 & 0.688 & \underline{0.719} & 0.654 & 0.688 & 0.644 & 0.713 & 0.675 & \textbf{0.762} & 0.694 \\
& \mbox{mF1 $\uparrow$} & 0.633 & 0.789 & \underline{0.815} & 0.755 & 0.788 & 0.744 & 0.811 & 0.776 & \textbf{0.850} & 0.794 \\
\arrayrulecolor{black!25}\cmidrule[0.25pt]{1-12}\arrayrulecolor{black}

\multirow{3}{*}{Ice (LoRA)} & \mbox{RMSE $\downarrow$} & 0.043 & 0.050 & 0.086 & 0.049 & 0.045 & 0.054 & 0.042 & 0.047 & \underline{0.033} & \textbf{0.031} \\
& \mbox{MAE $\downarrow$} & 0.029 & 0.034 & 0.062 & 0.034 & 0.031 & 0.038 & 0.029 & 0.032 & \underline{0.023} & \textbf{0.021} \\
& \mbox{$R^2$ $\uparrow$} & 0.975 & 0.967 & 0.900 & 0.968 & 0.973 & 0.961 & 0.976 & 0.970 & \underline{0.985} & \textbf{0.987} \\
\bottomrule
\end{tabular}

\par\vspace{2pt}
\begin{minipage}{\textwidth}
\textit{Notes.}\ Encoder parameters (millions; including input modules, excluding decoders and task heads), left to right: 91.2, 86.6, 111.4, 90.5, 90.6, 85.6, 195.1, 90.8, 204.3/199.8 (Moonstone/SomBench), and 91.2. mF1: macro-averaged F1; mAP: mean average precision; MAE: mean absolute error. IMP metrics average both classes. m$R^2$: channel-macro-averaged $R^2$; mRMSE uses training-standardized targets.
\end{minipage}
\endgroup
\end{table*}

\paragraph{Results on Moonstone Benchmark.} We evaluate all six tasks with frozen-encoder probing (LP) and full fine-tuning (FFT). Table~\ref{tab:benchmark_combined} compares models under the indicated adaptation setting for each task, with the same setting applied to all models. Complete LP and FFT results are provided in the supplementary material.

With the encoder frozen, Luniris leads in geological-unit classification, cross-modal thermal prediction, and mare segmentation, supporting both semantic discrimination and thermal-property estimation from non-thermal observations. Geological classification accuracy exceeds the strongest evaluated baseline by 4.41 percentage points.

\begin{figure*}[t]
    \centering
    \includegraphics[width=\textwidth]{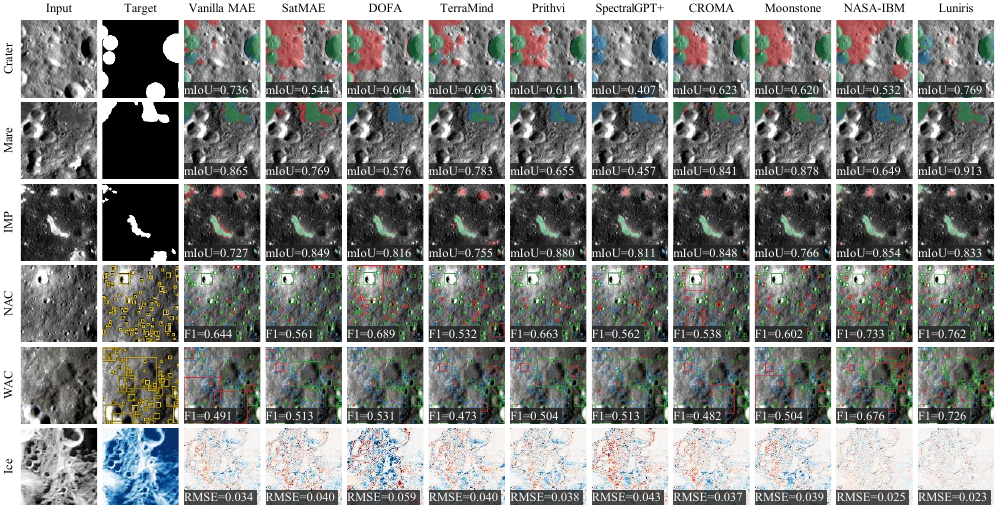}
    \caption{Qualitative comparison on Crater, Mare, IMP, NAC, WAC, and Ice (top to bottom). Models share the same scene for each task. Green/red/blue denote true/false positives and false negatives; dashed boxes indicate missed detections. Ice shows prediction-minus-target residuals: red indicates overestimation and blue underestimation. Scores are scene-level metrics.}
    \label{fig:downstream}
\end{figure*}

Full fine-tuning also yields leading surface-age classification accuracy and crater segmentation results, with crater mean intersection over union (mIoU) improving by 2.76 percentage points over the strongest evaluated baseline. Figure~\ref{fig:downstream} (top) illustrates the agreement between predicted and annotated crater outlines and mare regions.

Under the reported settings, Luniris also outperforms Vanilla MAE pretrained on the same corpus on five tasks, supporting the benefits of its pretraining design under matched data.

\paragraph{Results on SomBench.} SomBench tests transfer to datasets not used in Luniris pretraining through WAC and Narrow Angle Camera (NAC) crater detection, irregular mare patch (IMP) segmentation, and ice prospectivity regression. WAC uses five-band \(512\times512\) crops, whereas NAC and IMP use finer-resolution, single-band \(256\times256\) crops. All three use \(8\times8\) patches, compared with \(16\times16\) during pretraining. Ice uses nine channels across eight modalities. We evaluate all four tasks with both low-rank adaptation (LoRA) and full fine-tuning. Complete LoRA and FFT results are provided in the appendix.

Luniris achieves strong crater-detection performance on WAC and NAC and competitive ice prospectivity estimation under the reported adaptation settings. Figure~\ref{fig:downstream} illustrates crater localization across WAC and NAC observation scales. Under LoRA, Luniris reduces Ice root mean squared error (RMSE) by 5.1\% relative to NASA--IBM Lunar FM, extending transfer evidence to physical-property estimation.

Compared with Vanilla MAE pretrained on the same corpus, Luniris improves all four tasks under the reported settings, including IMP mIoU from 0.553 to 0.694. Together, these results support transfer through downstream adaptation across changes in spatial resolution, modality availability, and input organization.

\begin{figure*}[t]
    \centering
    \includegraphics[width=\textwidth]{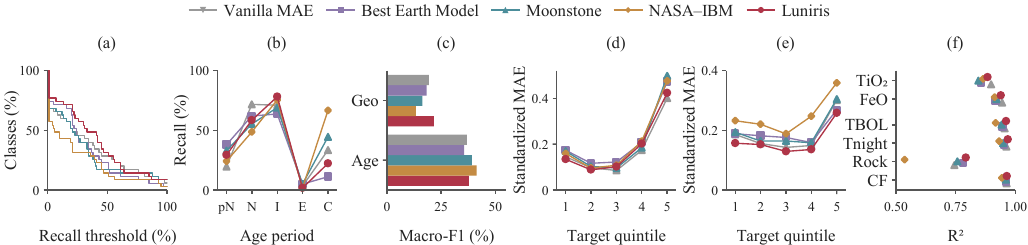}
    \caption{Moonstone prediction analysis: (a) Geo recall coverage; (b) Age-class recall; (c) macro-F1; (d,e) Comp/Therm MAE across test-target quintiles; (f) per-target $R^2$. MAE is divided by training-set standard deviations and averaged across channels. pN/N/I/E/C: pre-Nectarian/Nectarian/Imbrian/Eratosthenian/Copernican; TBOL/Tnight: bolometric/nighttime temperature; Rock: rock abundance; CF: Christiansen feature. Best Earth: SatMAE-B/DOFA-B for Geo/Age by accuracy; SpectralGPT+/CROMA-B for Comp/Therm by m$R^2$.}
    \label{fig:downstream_analysis}
\end{figure*}

\paragraph{Downstream Prediction Analysis.} Figure~\ref{fig:downstream_analysis} shows broader geological-class recognition: more Geo classes exceed intermediate recall thresholds, consistent with Luniris's higher macro-F1. Its Age predictions also achieve higher recall for the Imbrian class than the compared baselines, revealing a specific strength in geological-period discrimination.

The examples in Figure~\ref{fig:downstream} illustrate improved target delineation and localization. In Crater, Luniris suppresses broad false-positive regions connecting adjacent craters in Moonstone and NASA--IBM predictions. In Mare, it recovers foreground extensions partially missed by Moonstone. For NAC, it recovers small craters missed by Moonstone while producing fewer spurious detections than NASA--IBM. In WAC, NASA--IBM and Luniris recover more annotated craters than the Earth-pretrained baselines; Luniris additionally avoids an oversized false-positive box in NASA--IBM's prediction while retaining large shadowed craters.

For regression, Luniris maintains lower thermal prediction errors across all target-value quintiles and improves rock-abundance estimation over both lunar baselines (Figure~\ref{fig:downstream_analysis}). It also reduces composition errors in the highest-value quintile relative to Moonstone and NASA--IBM. In the Ice example (Figure~\ref{fig:downstream}), Luniris reduces spatially structured residuals relative to Moonstone, more closely recovering the target field's spatial variation.

\subsection{Ablation Studies}
\label{sec:ablation}

\begin{table}[t]
\centering
\caption{Ablation of hierarchical support embeddings (E) and RSR (R). $\checkmark/\times$: enabled/disabled. Settings and metrics follow Table~\ref{tab:benchmark_combined}. Bold/underlined: best/second best, ranked before rounding.}
\label{tab:ablation}

\begingroup
\newcommand{\ablationOn}{\textcolor{green!45!black}{\ensuremath{\checkmark}}}
\newcommand{\ablationOff}{\textcolor{red!75!black}{\ensuremath{\times}}}
\small
\setlength{\tabcolsep}{1.5pt}
\setlength{\aboverulesep}{0.6pt}
\setlength{\belowrulesep}{0.6pt}
\renewcommand{\arraystretch}{0.95}

\begin{tabular*}{\columnwidth}{@{\extracolsep{\fill}}lccccccc@{}}
\toprule
\multirow{2}{*}{Model}
& \multirow{2}{*}{E}
& \multirow{2}{*}{R}
& Geo & Therm & Crater & NAC & Ice \\
& & &
\mbox{Acc $\uparrow$} &
\mbox{m$R^2$ $\uparrow$} &
\mbox{mIoU $\uparrow$} &
\mbox{mAP $\uparrow$} &
\mbox{RMSE $\downarrow$} \\
\midrule

Vanilla MAE & -- & -- & 0.335 & 0.907 & 0.656 & 0.081 & 0.043 \\
\cmidrule[0.25pt]{1-8}

Core & \ablationOff & \ablationOff & 0.372 & \underline{0.947} & 0.678 & 0.156 & \underline{0.032} \\
w/o E & \ablationOff & \ablationOn & 0.374 & \textbf{0.954} & \underline{0.680} & 0.153 & 0.032 \\
w/o R & \ablationOn & \ablationOff & \underline{0.378} & 0.941 & 0.647 & \underline{0.164} & 0.032 \\
\cmidrule[0.25pt]{1-8}

\textbf{Luniris} & \ablationOn & \ablationOn & \textbf{0.379} & 0.923 & \textbf{0.693} & \textbf{0.165} & \textbf{0.031} \\
\bottomrule
\end{tabular*}
\endgroup
\end{table}

Table~\ref{tab:ablation} evaluates all four combinations of hierarchical support embeddings ($E$) and RSR ($R$). Core disables both components. Removing $E$ retains support normalization, attention-key eligibility, and valid-target selection, evaluating coverage embeddings jointly within the support-conditioned framework.

Core outperforms the same-data Vanilla MAE control on all five reported tasks, raising Crater mIoU from 0.656 to 0.678 and NAC mAP from 0.081 to 0.156. The component ablations therefore measure the additional contributions of support embeddings and resolution-supported reconstruction on top of this base formulation.

Support embeddings improve geological classification accuracy and NAC detection mAP both with and without RSR, supporting the value of encoding coverage alongside support-based validity handling. The effect differs for thermal prediction: the variant with $R$ and without $E$ achieves the highest macro-$R^2$ of 0.9537, versus 0.9230 for Luniris.

The joint configuration leads on four of the five reported tasks. On Crater, the ablations reveal an interaction between coverage encoding and auxiliary reconstruction: adding RSR increases mIoU from 0.6474 to 0.6931 with $E$, a gain of 4.57 percentage points, compared with an increase from 0.6783 to 0.6795 without $E$, a gain of 0.12 points. Luniris also exceeds Core by 1.48 percentage points. This supports complementary component effects on Crater.
\section{Conclusion}
\label{sec:conclusion}
We propose Luniris, a lunar foundation model that incorporates observation support and effective spatial resolution into multimodal masked pretraining. Hierarchical support conditions encoding and reconstruction target selection, while RSR adapts auxiliary supervision scales to targets and valid visible observations within the same spatial patch. Experiments on Moonstone Benchmark and SomBench demonstrate competitive performance across ten downstream tasks spanning classification, regression, segmentation, and object detection. Matched-data comparisons and component ablations support the contribution of the pretraining design, while SomBench results demonstrate transfer across changes in modality availability, spatial resolution, and input organization. Future work will extend this formulation to jointly model observations from different instruments at their native spatial resolutions.
\section*{LLM Usage Statement}

OpenAI Codex was used to assist with drafting and revising the manuscript. The authors retain full responsibility for the final manuscript, including the accuracy of its content.
\FloatBarrier
{
    \small
    \bibliographystyle{ieeenat_fullname}
    \bibliography{main}
}

\clearpage
\maketitlesupplementary
\appendix
\setcounter{equation}{0}
\setcounter{table}{0}
\setcounter{figure}{0}
\renewcommand{\theequation}{S\arabic{equation}}
\renewcommand{\thetable}{S\arabic{table}}
\renewcommand{\thefigure}{S\arabic{figure}}
\renewcommand{\theHequation}{supp.\arabic{equation}}
\renewcommand{\theHtable}{supp.\arabic{table}}
\renewcommand{\theHfigure}{supp.\arabic{figure}}
% This supplement is appended to main.tex so that references to the main paper resolve.
% Implementation basis: CUBE_MAE_METHOD_PACKAGE_20261026; main MSE/150-epoch configuration.
% Terminology follows the current Method: lattice, spatial patch, crop, Evidence, and RSR.
% Engineering filenames retain their original names.
% Sources: modality_groups.py, cube_mae.py, tube_mask.py, rscr_cube.py,
% physical_constraints.py, cube_mae_runner.py, cube_mae_vitb_train.yaml,
% spherical_center_sampler.py, geo.py, destination_resampler.py, frozen_normalizer.py.

\makeatletter
\setlength{\@dblfptop}{0pt}
\setlength{\@dblfpsep}{12pt}
\setlength{\@dblfpbot}{0pt plus 1fil}
\makeatother

\section{Additional Method Details}
\label{sec:supp_method}

This supplement explains how Luniris inputs are prepared, how masking and Evidence are applied, and how the reconstruction losses are computed. It also gives the training configuration. Terms and symbols follow \cref{sec:method}. A lattice contains all channels of one modality within one spatial patch; a crop is one input image. The index $b$ identifies a crop in the minibatch, $p$ identifies a pixel in that crop, and $(i(p),j(p))$ identifies the patch containing that pixel.

\subsection{Input Construction and Geographic Organization}
\label{sec:supp_input}

\paragraph{Spherical Sampling and Partitioning.}
On a longitude--latitude grid, each pixel covers less lunar surface near the poles than near the equator. Spherical Sampling and Partitioning (SSP) samples crop centers uniformly over the lunar surface. Longitude $\ell$ and sine latitude $z$ are drawn uniformly:
\begin{equation}
\ell\sim\mathcal{U}[-\pi,\pi),\qquad z\sim\mathcal{U}[-1,1],\qquad \varphi=\arcsin z.
\label{eq:supp_spherical_sampling}
\end{equation}
Here $\varphi$ is latitude. HEALPix divides the lunar surface into 768 equal-area regions at $N_{\mathrm{side}}=8$. A fixed hash ordering assigns 614 regions to training, 77 to validation, and 77 to testing. A candidate crop is accepted only if its center, four corners, and four edge midpoints all belong to the requested split. The corners and edge midpoints are evaluated on the destination pixel-center grid. Source pixels are checked for validity separately during resampling. Pretraining and the Moonstone downstream benchmark use the same regional split. SomBench uses its official datasets and splits.

\paragraph{Local metric crops.}
A $256\times256$ grid is placed around each crop center. Grid positions are defined by eastward and northward distances from the center and mapped onto the lunar sphere. The spacing is $\delta\approx236.901175$ metres per pixel, giving a nominal crop width of $256\delta\approx60.647$ km. The same spacing is used at every latitude. Each source product is resampled onto this grid before its channels are normalized.

\paragraph{Values and observation support.}
Bilinear resampling combines valid source pixels near each destination pixel $p$. For source neighbor $a$, $\omega_{pa}$ is its interpolation weight, $\xi_a$ is its measured value, and $\nu_a$ is one if the source pixel is valid and zero otherwise. Omitting channel indices, the observation support and resampled value are
\begin{equation}
\begin{aligned}
V(p)&=\sum_a\omega_{pa}\nu_a,\\
X^{\mathrm{phys}}(p)&=\frac{\sum_a\omega_{pa}\nu_a\xi_a}{V(p)},\qquad V(p)>0.
\end{aligned}
\label{eq:supp_resampling}
\end{equation}
Invalid source values are set to zero before multiplication. A destination pixel is also set to zero if none of its source neighbors is valid. The support value $V$ records how much interpolation weight comes from valid neighbors: $V=1$ means full support, $0<V<1$ partial support, and $V=0$ no support. These continuous values weight token content and are averaged to obtain the three Evidence summaries. A pixel can serve as a reconstruction target whenever $V>0$. Source validity comes from the raster masks, validity maps, or observation counts supplied for each product. Stokes validity is checked jointly across its four channels.

\paragraph{Frozen normalization.}
Each channel uses a fixed transform $\mathcal{T}_{m,c}$, followed by normalization with a mean $\mu_{m,c}^{\mathrm{norm}}$ and standard deviation $\sigma_{m,c}^{\mathrm{norm}}$ estimated from training regions. Values with positive support are normalized as
\begin{equation}
X_{m,c}=\frac{\mathcal{T}_{m,c}(X^{\mathrm{phys}}_{m,c})-\mu_{m,c}^{\mathrm{norm}}}{\sigma_{m,c}^{\mathrm{norm}}}.
\label{eq:supp_normalization}
\end{equation}
The transform is either identity or $\mathcal{T}_{m,c}(x)=\operatorname{asinh}(x/\kappa_{m,c})$, as listed in \cref{tab:supp_modalities}, with $\kappa_{m,c}=1$ for all asinh-transformed channels. Normalization parameters remain fixed across crops and data splits, and missing values remain zero. Before physical regularization, predictions and targets are converted back to physical units:
\begin{equation}
X^{\mathrm{phys}}_{m,c}=\mathcal{T}_{m,c}^{-1}\!\left(\sigma_{m,c}^{\mathrm{norm}}X_{m,c}+\mu_{m,c}^{\mathrm{norm}}\right).
\label{eq:supp_inverse_normalization}
\end{equation}

\subsection{Modality--Spatial Lattices and Evidence}
\label{sec:supp_evidence}

\paragraph{Channel organization.}
The 30 input channels form seven physical modalities, listed in \cref{tab:supp_modalities}. Modality-grouped tokenization follows the organization used in Moonstone~\cite{prasad2026moonstone}. Each modality has a dedicated convolutional tokenizer $\operatorname{Conv}_m$ and reconstruction head. A lattice contains all channels of that modality within a single spatial patch.

\begin{table}[t]
\centering
\small
\setlength{\tabcolsep}{2pt}
\renewcommand{\arraystretch}{1.08}
\caption{Modality configuration in token order. Channel indices are zero-based. Composition includes Clementine (27) and Lunar Prospector Gamma Ray Spectrometer (LP-GRS) (28--29). Gravity and LP-GRS use configured scale proxies. Transforms precede standardization.}
\label{tab:supp_modalities}
\begin{tabular*}{\linewidth}{@{\extracolsep{\fill}}lcrl@{}}
\toprule
Modality & Channels & $g$ (m/pixel) & Transform\\
\midrule
Composition  & 27     & 236.901 & asinh\\
             & 28--29 & 15,000  & identity\\
\addlinespace[2pt]
Gravity      & 16--18 & 5,000   & identity\\
Hapke        & 23--26 & 398.991 & identity\\
Radar Stokes & 19--22 & 236.901 & asinh\\
M$^3$ reflectance & 8--15 & 236.901 & asinh\\
Surface      & 0--3   & 236.901 & identity\\
Thermal      & 4--7   & 236.901 & identity\\
\bottomrule
\end{tabular*}
\end{table}

\paragraph{Lattice masking.}
With patch size $P=16$, each crop contains $16\times16$ spatial patches and $N=G H_cW_c=1792$ lattices. The set $\mathcal{I}=\{(m,i,j)\}$ indexes these lattices. For each crop $b$, each lattice receives an independent uniform random score. Sorting these scores gives a random order $\pi_b$. The first $N_{\mathrm{vis}}=\lfloor(1-\rho)N\rfloor$ lattices remain unmasked:
\begin{equation}
\begin{aligned}
\mathcal{O}_b&=\{\pi_{b,1},\ldots,\pi_{b,N_{\mathrm{vis}}}\},\\
\mathcal{M}_b&=\mathcal{I}\setminus\mathcal{O}_b,\\
M_{b,m,i,j}&=\mathbf{1}[(m,i,j)\in\mathcal{M}_b].
\end{aligned}
\label{eq:supp_masking}
\end{equation}
At $\rho=0.75$, each crop has 448 unmasked lattices and 1344 masked lattices. Only the total count is fixed; the unmasked count can differ across modalities and spatial patches. Mask sampling does not depend on observation values or support. All channels and pixels within a lattice share its mask.

\paragraph{Support summaries and attention keys.}
The three summaries $v^{\mathrm{ch}}$, $v^{\mathrm{lat}}$, and $v^{\mathrm{crop}}$ in \cref{eq:evidence_support} are computed before artificial masking. They remain available for both masked and unmasked lattices. An unmasked lattice can serve as an attention key only if it contains at least one pixel that is valid in every channel:
\begin{equation}
\begin{aligned}
J_{b,m,i,j,u,v}&=\prod_{c=1}^{C_m}\mathbf{1}[V_{b,m,i,j,c,u,v}>0],\\
K^{\mathrm{obs}}_{b,m,i,j}&=\mathbf{1}\!\left[\sum_{u,v}J_{b,m,i,j,u,v}>0\right].
\end{aligned}
\label{eq:supp_joint_support}
\end{equation}
Here $J=1$ marks a pixel that is valid in every channel, and $K^{\mathrm{obs}}=1$ allows the unmasked lattice to serve as an attention key. The average support $v^{\mathrm{lat}}$ alone cannot establish whether the channels share a valid pixel. Within each gravity, radar Stokes, M3 reflectance, or surface-property modality, channels have the same valid-pixel mask. This agreement is checked at input. Composition, Hapke, and thermal channels can have different valid-pixel masks.

\paragraph{Token content and Evidence embeddings.}
Each $\operatorname{Conv}_m$ uses kernel size and stride $P$. Input values with no support are first set to zero. The input is then multiplied by $V$ and passed through $\operatorname{Conv}_m$. Its output is divided by $\max(v_{m,i,j}^{\mathrm{lat}},\epsilon)$, where $\epsilon=10^{-4}$ prevents division by a very small value. If the lattice has no support, its content representation is set to zero. Modality, position, and Evidence embeddings are then added as in \cref{eq:evidence_tokens}.

The Evidence embedding $E_{m,i,j}^{r}(V)$ is the sum of three projected support summaries. The encoder ($r=\mathrm{enc}$) and decoder ($r=\mathrm{dec}$) have separate projection parameters. Each modality has its own channel projection $\mathcal{P}_{m,r}^{\mathrm{ch}}$. Within each branch, all modalities share the lattice projection $\mathcal{P}_{r}^{\mathrm{lat}}$ and the crop projection $\mathcal{P}_{r}^{\mathrm{crop}}$. The resulting crop embedding is added at every spatial patch of that modality. Since lattice-level support is the mean of channel-level support, its linear embedding can be absorbed into the channel-level projection. We retain this decomposition as a parameterization of local support, alongside crop-level support. The $E$ ablation evaluates these embeddings jointly rather than isolating the contribution of each level.

\paragraph{Encoding and decoder restoration.}
The encoder receives only unmasked tokens. Their $K^{\mathrm{obs}}$ values form the mask $K_{\mathcal{O}}(V)$ in \cref{eq:luniris_encoding}. A token with $K^{\mathrm{obs}}=0$ cannot be used as a key or value, but it still acts as a query. If every unmasked token has $K^{\mathrm{obs}}=0$, the first token is allowed as a key so that the attention softmax has at least one available entry. This numerical fallback does not change which pixels contribute to the loss.

The encoder output $Z$ is projected to the decoder's embedding dimension. Learned mask tokens fill the masked lattice positions, restoring the original lattice order. Decoder modality, position, and Evidence embeddings are then added. The decoder uses the following key-validity mask, where one means that a token can serve as a key:
\begin{equation}
K^{\mathrm{dec}}_{b,m,i,j}=M_{b,m,i,j}+(1-M_{b,m,i,j})K^{\mathrm{obs}}_{b,m,i,j}.
\label{eq:supp_decoder_keys}
\end{equation}
Mask tokens remain attention keys, while unmasked tokens follow the same observation-support rule as in the encoder. Each modality-specific head predicts $C_mP^2$ values for each masked lattice.

\paragraph{Pixel-level supervision.}
Every pixel inherits its lattice's artificial mask: $M_{b,m}(p)=M_{b,m,i(p),j(p)}$. A pixel is used as a reconstruction target when the lattice is masked and the target channel has positive observation support:
\begin{equation}
\begin{aligned}
T_{b,m,c}(p)&=M_{b,m}(p)\,\mathbf{1}[V_{b,m,c}(p)>0],\\
T^{\cap}_{b,m}(p)&=M_{b,m}(p)\prod_{c=1}^{C_m}\mathbf{1}[V_{b,m,c}(p)>0].
\end{aligned}
\label{eq:supp_target_masks}
\end{equation}
Base reconstruction and RSR select target pixels separately for each channel using $T$. Spectral regularization uses a pixel only when $T=1$ in both adjacent bands. Stokes regularization requires all four channels to be valid at that pixel, as specified by $T^{\cap}$. Unmasked positions in the full prediction maps contain placeholder values and are excluded from the reconstruction losses.

\subsection{Resolution-Supported Reconstruction}
\label{sec:supp_rscr}

\paragraph{Effective resolution.}
The registered source resolution $g^{\mathrm{src}}_{m,c}$ describes the spatial scale supported by each channel. It is taken from the native resolution or estimated for the source product. The effective resolution after alignment is the coarser of this source resolution and the shared grid spacing:
\begin{equation}
g_{m,c}=\max(g^{\mathrm{src}}_{m,c},\delta).
\label{eq:supp_resolution_floor}
\end{equation}
All resolutions and $\delta$ are measured in metres per pixel, so a larger value means a coarser resolution. If the source is coarser than the shared grid, $g_{m,c}$ retains the source resolution. If the source is finer, $g_{m,c}$ uses the grid spacing $\delta$. These values determine the RSR supervision scales.

\paragraph{Valid unmasked channels.}
For a masked lattice, RSR considers only unmasked lattices within the same spatial patch. A channel is available there if it contains at least one supported pixel:
\begin{equation}
h_{b,m,c,i,j}=\mathbf{1}\!\left[\sum_{u,v}\mathbf{1}[V_{b,m,i,j,c,u,v}>0]>0\right].
\label{eq:supp_channel_presence}
\end{equation}
For a modality whose channels share a valid-pixel mask, all channels must contain valid pixels in the patch. This gives $\widetilde h_{b,m,c,i,j}=\prod_{c'=1}^{C_m}h_{b,m,c',i,j}$. For modalities with separate channel masks, each channel is checked on its own, so $\widetilde h=h$. Channels that pass this check and belong to unmasked lattices form the set
\begin{equation}
\mathcal{C}_{b,i,j}=\{(n,c'):M_{b,n,i,j}=0,\ \widetilde h_{b,n,c',i,j}=1\}.
\label{eq:supp_visible_set}
\end{equation}
This is the set $\mathcal{C}_{i,j}$ from the main paper, with $b$ added to identify the crop. RSR uses the resolutions of channels in this set. Membership in $\mathcal{C}_{b,i,j}$ is determined by channel-level observation availability and does not require the parent token to satisfy $K^{\mathrm{obs}}=1$. The set therefore describes observed channels used for scale selection, while attention-key eligibility is determined separately by \cref{eq:supp_joint_support}.

When $\mathcal{C}_{b,i,j}$ is nonempty, RSR finds the finest resolution among these unmasked channels. It then compares that resolution with each masked target channel and selects the coarser one:
\begin{equation}
\begin{aligned}
g^{\mathrm{vis}}_{b,i,j}&=\min_{(n,c')\in\mathcal{C}_{b,i,j}}g_{n,c'},\\
g^{\mathrm{sup}}_{b,m,c,i,j}&=\max(g_{m,c},g^{\mathrm{vis}}_{b,i,j}).
\end{aligned}
\label{eq:supp_supported_scale}
\end{equation}
If the finest available observation is at least as fine as the target channel, RSR uses the target resolution. If it is coarser, RSR uses that coarser resolution to set the pooling grid and compare low-frequency content. When no valid unmasked channel is available in the patch, RSR is skipped. Valid masked pixels in that patch still contribute to the base reconstruction loss.

\paragraph{Pooling regions.}
Let $s=(b,m,c,i,j)$ identify one target channel in a masked lattice. The selected scale determines the pooling factor and output dimensions:
\begin{equation}
\begin{aligned}
k_s&=\operatorname{clip}\!\left(\operatorname{round}(g^{\mathrm{sup}}_s/\delta),1,\min(H,W)\right),\\
n_y&=\lceil H/k_s\rceil,\qquad n_x=\lceil W/k_s\rceil.
\end{aligned}
\label{eq:supp_pooling_factor}
\end{equation}
Here $g^{\mathrm{sup}}_s=g^{\mathrm{sup}}_{b,m,c,i,j}$. Adaptive average pooling first sets the number of output regions to $n_y\times n_x$. It then calculates which input pixels belong to each region. For the region at output row $a$ and column $d$, the input row and column indices are
\begin{equation}
\begin{aligned}
\mathcal{R}^{y}_a&=\{\lfloor aH/n_y\rfloor,\ldots,\lceil(a+1)H/n_y\rceil-1\},\\
\mathcal{R}^{x}_d&=\{\lfloor dW/n_x\rfloor,\ldots,\lceil(d+1)W/n_x\rceil-1\},
\end{aligned}
\label{eq:supp_pooling_bins}
\end{equation}
where $0\leq a<n_y$ and $0\leq d<n_x$. The region contains every combination of these row and column indices. Together, the regions cover the full crop. Neighboring regions can overlap when the input size is not divisible by the output size. For example, a $256\times256$ crop with $k_s=63$ produces a $5\times5$ output grid with pooling regions of $52\times52$ pixels. Each regional mean is subsequently restricted to valid pixels inside the target patch.

\paragraph{Averaging within the target patch.}
Within each pooling region, the mean uses only valid target pixels inside the target patch $(i,j)$. Let $\mathcal{R}_q$ denote the pooling region at $q=(a,d)$. The pixels used in the mean are
\begin{equation}
\Omega_{s,q}=\{p\in\mathcal{R}_q:(i(p),j(p))=(i,j),\ V_{b,m,c}(p)>0\}.
\label{eq:supp_pooling_support}
\end{equation}
The set $\Omega_{s,q}$ is $\Omega_q$ from the main paper, with $s$ added to identify the target. If $U$ is the prediction or target channel map, its mean over these pixels is
\begin{equation}
\Pi_s(U;V)_q=\frac{1}{|\Omega_{s,q}|}\sum_{p\in\Omega_{s,q}}U(p).
\label{eq:supp_cell_projection}
\end{equation}
Prediction and target means use exactly the same pixels. Every included pixel has equal weight, whether its support is partial or full. Missing pixels and pixels outside the target patch are excluded. Thus, the selected resolution determines the pooling grid, and each mean uses the valid part of the target patch within that grid region. Regions containing no valid target pixels are skipped. The remaining regions form $\mathcal{Q}_s=\{q:|\Omega_{s,q}|>0\}$.

The implementation first sets excluded pixels to zero and pools the channel values. It applies the same pooling operation to a binary mask that is one at included pixels and zero elsewhere. Dividing the first result by the second gives the mean over included pixels. The denominator is bounded below by machine epsilon to avoid numerical problems. Regions with a zero pooled mask are excluded from the loss.

\paragraph{RSR loss.}
Let $\hat{\mu}_{s,q}=\Pi_s(\widehat X_s;V)_q$ be the prediction mean and $\mu_{s,q}=\Pi_s(X_s;V)_q$ the target mean in region $q$. RSR computes their squared difference. The loss first averages these errors over valid regions within each target, then equally over targets:
\begin{equation}
\mathcal{L}_{\mathrm{RSR}}=\frac{1}{|\mathcal{S}|}\sum_{s\in\mathcal{S}}\frac{1}{|\mathcal{Q}_s|}\sum_{q\in\mathcal{Q}_s}(\hat{\mu}_{s,q}-\mu_{s,q})^2.
\label{eq:supp_rscr_loss}
\end{equation}
A target belongs to $\mathcal{S}$ if it has at least one valid masked pixel and its patch has an available unmasked channel, meaning $\mathcal{C}_{b,i,j}$ is nonempty. If there are no such targets, the RSR loss is zero. After averaging its regional errors, each target channel in each masked lattice receives equal weight across the minibatch. RSR is added to the base pixel-level loss. The base loss uses the separate modality-balanced averaging described below.

\paragraph{RSR support and context statistics.} We audit 512 training crops using the training resolution registry, observation support maps, and masking procedure. Of 2,755,914 candidate target channel--patch pairs with valid masked pixels, 19.65\% lack valid unmasked context in the same patch and are excluded from RSR. Among the remaining 2,214,460 pairs, pooling factors $k\in\{1,2,21,63\}$ occur with frequencies of 56.25\%, 19.23\%, 17.41\%, and 7.11\%, respectively, counting each eligible pair equally; 24.51\% have $k>P$, where $P=16$. The corresponding median numbers of valid pixels per nonempty pooling region are 1, 4, 70, and 160, with each mean restricted to the target patch of nominal side length 3.79\,km. Separately, 1.22\% of context-channel entries belong to tokens with $K^{\mathrm{obs}}=0$, appearing in the context sets of 3.44\% of eligible targets. All observed mismatches arise from partial channel missingness, predominantly in the thermal modality. Excluding these entries in an offline diagnostic changes the pooling grid for 0.415\% of all eligible targets and leaves 0.940\% without same-patch context. The reported training results retain the original channel-availability rule.

\subsection{Complete Pretraining Objective}
\label{sec:supp_objective}

\paragraph{Modality-balanced reconstruction.}
The base reconstruction loss gives equal weight to each modality after averaging its channel losses. The set $\mathcal{P}_{m,c}=\{(b,p):T_{b,m,c}(p)=1\}$ contains the valid masked pixels of channel $c$ within each local microbatch. Channels with at least one such pixel form $\mathcal{C}_m^+$. Modalities with at least one such channel form $\mathcal{G}^+$. Squared errors are computed in normalized input space:
\begin{equation}
\begin{aligned}
\ell_{m,c}&=\frac{1}{|\mathcal{P}_{m,c}|}\sum_{(b,p)\in\mathcal{P}_{m,c}}\bigl[\widehat X_{b,m,c}(p)-X_{b,m,c}(p)\bigr]^2,\\
\mathcal{L}_{\mathrm{rec}}&=\frac{1}{|\mathcal{G}^+|}\sum_{m\in\mathcal{G}^+}\frac{1}{|\mathcal{C}_m^+|}\sum_{c\in\mathcal{C}_m^+}\ell_{m,c}.
\end{aligned}
\label{eq:supp_balanced_mse}
\end{equation}
The three averages are taken over pixels within each channel, channels within each modality, and modalities. Channels and modalities with no valid targets are skipped. During distributed training, each GPU computes this loss on its local microbatch. Gradients are averaged equally across training processes and accumulation steps.

\paragraph{Spectral slope matching.}
The eight M3 bands have wavelengths $(750,950,1000,1250,1580,2000,2817,2857)$ nm. After inverse normalization, let $R_c$ denote reflectance at wavelength $\lambda_c$. The slope between adjacent bands is
\begin{equation}
D_cR=\frac{R_{c+1}-R_c}{\lambda_{c+1}-\lambda_c},\qquad c=1,\ldots,7.
\label{eq:supp_spectral_slope}
\end{equation}
An adjacent band pair contributes to the loss only at pixels where both bands are valid masked targets. These pixel locations, indexed by crop and pixel $(b,p)$, form $\mathcal{U}_c$. Band pairs with at least one such location form $\mathcal{C}_{\mathrm{spec}}^+$. The spectral loss compares predicted and observed slopes:
\begin{equation}
\mathcal{L}_{\mathrm{spec}}=\frac{1}{|\mathcal{C}_{\mathrm{spec}}^+|}\sum_{c\in\mathcal{C}_{\mathrm{spec}}^+}\frac{1}{|\mathcal{U}_c|}\sum_{(b,p)\in\mathcal{U}_c}\bigl[D_c\widehat R_b(p)-D_cR_b(p)\bigr]^2.
\label{eq:supp_spectral_loss}
\end{equation}
For each adjacent band pair, the slope error is averaged over its valid pixels. These pair losses are then averaged equally. Dividing by the wavelength difference accounts for the unequal spacing between M3 bands.

\paragraph{Stokes regularization.}
The Stokes term penalizes negative intensity and a polarization magnitude larger than intensity. A pixel contributes only if its radar lattice is masked and all four Stokes channels are valid at that pixel. These crop--pixel locations form $\mathcal{U}_{\mathrm{St}}$. After predictions are converted back to physical units, the two penalties are based on
\begin{equation}
\begin{aligned}
\zeta&=\sqrt{\widehat S_2^2+\widehat S_3^2+\widehat S_4^2+10^{-12}},\\
r_I&=\max(0,-\widehat S_1),\qquad r_P=\max(0,\zeta-\widehat S_1).
\end{aligned}
\label{eq:supp_stokes_residuals}
\end{equation}
Here $\widehat S_1$ is the predicted intensity, and $\zeta$ is the magnitude of the predicted polarization vector $(\widehat S_2,\widehat S_3,\widehat S_4)$ with a small constant added inside the square root for stability. The value $r_I$ measures how far intensity falls below zero; $r_P$ measures how far polarization magnitude exceeds intensity. Their squared values give the loss
\begin{equation}
\mathcal{L}_{\mathrm{St}}=\frac{1}{|\mathcal{U}_{\mathrm{St}}|}\sum_{(b,p)\in\mathcal{U}_{\mathrm{St}}}\bigl[r_I(b,p)^2+r_P(b,p)^2\bigr].
\label{eq:supp_stokes_loss}
\end{equation}
The spectral and Stokes terms are each zero when their valid target sets are empty.

\paragraph{Combined loss and physical warmup.}
The objective in \cref{eq:pretraining_objective} uses $\lambda_{\mathrm{RSR}}=0.25$, $\lambda_{\mathrm{phys}}=7.5$, and equal weights for the two physical terms:
\begin{equation}
\begin{aligned}
\mathcal{L}_{\mathrm{total}}&=\mathcal{L}_{\mathrm{rec}}+0.25\mathcal{L}_{\mathrm{RSR}}\\
&\quad+7.5\alpha(t)(\mathcal{L}_{\mathrm{spec}}+\mathcal{L}_{\mathrm{St}}),\\
\alpha(t)&=\min(1,t/T_{\mathrm{phys}}),\qquad T_{\mathrm{phys}}=2500.
\end{aligned}
\label{eq:supp_total_loss}
\end{equation}
The physical regularization weight increases linearly until $t$ reaches $T_{\mathrm{phys}}$, then remains fixed. The index $t$ advances with each training loss evaluation. Specifically, let $A$ be the number of microbatches accumulated per optimizer update, $u\geq1$ the update index, and $a\in\{0,\ldots,A-1\}$ the microbatch index. The implementation uses $t=Au+a$. Physical warmup therefore advances for each microbatch, while the learning rate changes once per optimizer update.

\subsection{Architecture and Optimization Configuration}
\label{sec:supp_configuration}

All training runs use four NVIDIA H100 GPUs. The ViT-B encoder has 12 Transformer blocks, an embedding dimension of 768, and 12 attention heads. The decoder has 8 blocks, an embedding dimension of 512, and 16 heads. Both use layer normalization before attention and feed-forward layers, Gaussian error linear unit activations, and an MLP expansion ratio of 4. A final layer normalization follows each stack. Modality, position, and mask embeddings are learned. The encoder processes 448 unmasked tokens per crop. The decoder restores all 1792 lattice positions and uses modality-specific heads to reconstruct the masked lattices.

\begin{table}[t]
\centering
\caption{MSE/150-epoch pretraining configuration. One distributed training process runs on each GPU.}
\small
\begin{tabularx}{\columnwidth}{@{}>{\raggedright\arraybackslash}Xl@{}}
\toprule
Setting & Value\\
\midrule
GPUs & $4\times$ NVIDIA H100\\
Input size / patch size & $256\times256$ / $16\times16$\\
Encoder width / blocks / heads & 768 / 12 / 12\\
Decoder width / blocks / heads & 512 / 8 / 16\\
multilayer perceptron (MLP) expansion ratio & 4\\
Mask ratio & 0.75\\
Base reconstruction / RSR & MSE / MSE\\
Pretraining epochs & 150\\
Crops per epoch & 128,000\\
Microbatch per GPU & 8\\
Gradient accumulation & 2\\
Effective global batch & 64\\
Optimizer & AdamW\\
Adam coefficients & $(0.9,0.95)$\\
Weight decay & 0.05\\
Gradient-norm clipping & 1.0\\
Initial / peak / final learning rate & $2\!\times\!10^{-5}$ / $10^{-4}$ / $2\!\times\!10^{-6}$\\
learning-rate warmup & 500 optimizer updates\\
RSR / physical weight & 0.25 / 7.5\\
Physical warmup $T_{\mathrm{phys}}$ & 2500 loss evaluations\\
\bottomrule
\end{tabularx}

\label{tab:supp_configuration}
\end{table}

\paragraph{Learning rate and gradient accumulation.}
The effective batch size is $4\times8\times2=64$ crops, giving 2000 optimizer updates per epoch and 300,000 updates over 150 epochs. AdamW uses linear learning-rate warmup followed by cosine decay:
\begin{equation}
\eta(u)=\begin{cases}
\eta_0+(\eta_{\max}-\eta_0)u/U_w,&u\leq U_w,\\
\eta_{\min}+\dfrac{\eta_{\max}-\eta_{\min}}{2}[1+\cos(\pi r_u)],&u>U_w,
\end{cases}
\label{eq:supp_learning_rate}
\end{equation}
where $r_u=\min(1,(u-U_w)/(U_{\max}-U_w))$, $U_w=500$, and $U_{\max}=300000$. The initial, peak, and final learning rates are listed in \cref{tab:supp_configuration}. Each local microbatch loss is divided by $A=2$ before backpropagation. After accumulation, gradients are clipped to norm 1.0 before the optimizer update.

\section{Additional Data Details}
Luniris combines lunar morphology, thermal properties, reflectance, gravity, radar polarimetry, and composition. The resolution values used to configure RSR range from approximately 237\,m to 15\,km, despite a shared input grid. The training configuration assigns 5,000\,m to the gravity channels and 15,000\,m to the LP-GRS channels. This variation motivates the scale-dependent reconstruction supervision in RSR.

To provide four-component radar inputs on the same spatial grid as the other modalities, we constructed a Mini-RF Stokes mosaic from 6,832 S-band zoom-mode Level-2 scenes, retaining all four components ($S_1$--$S_4$). The scenes were mosaicked on a common lunar equirectangular grid at 1,024 pixels per degree (ppd; approximately 29.61 m per pixel at the equator), together with a valid-observation count map. Areas without coverage were retained as missing data. The mosaic was then downsampled to 128 ppd (approximately 236.9 m per pixel at the equator) to match the global grids of the other modalities. The four channels jointly cover 67.58\% of global grid pixels, corresponding to 61.51\% of the lunar surface after latitude weighting. This incomplete coverage motivates retaining explicit observation-validity information alongside the gridded values, as used in Luniris's Evidence conditioning.

For downstream evaluation, we retain Moonstone's task definitions while standardizing the geographic partitions across tasks and models. The released cross-modal thermal prediction script reports validation metrics without a separate test evaluation. Its spatial-block partitioning uses longitude--latitude grid cells whose surface areas vary with latitude, without an explicit geographic buffer between splits. We adopt SSP with equal-area HEALPix partition units and shared training/validation/test assignments. Geo, Age, Mare, and Crater each contain 10,556/550/931 samples, while target-availability filtering yields 10,548/549/931 for Comp and 7,352/444/419 for Therm. SomBench retains its official partitions, with test sets of 100 WAC images, 63 NAC images, 10 IMP tiles, and 25 Ice tiles.

\section{Additional Experimental Results}
\label{sec:supp_experiments}

We extend the main experiments with reconstructions across all 30 input channels and complete downstream comparisons on the Moonstone Benchmark and SomBench. These results provide additional evidence for reconstruction quality across physical modalities and downstream performance under different adaptation settings. LunarFM~\cite{girona2026lunarfm} is excluded from evaluation because we could not locate publicly accessible code or pretrained weights.

\paragraph{Model variants.} We use multispectral SatMAE-B, DOFA-B, TerraMind-v1-B, Prithvi-EO-2.0-100M-TL, SpectralGPT+, CROMA-B, and a reproduction of Moonstone MG-MAE. Vanilla MAE uses the same pretraining corpus as Luniris. The abbreviated table headers refer to these variants. Parameter counts include the encoder and input modules but exclude pretraining decoders and task heads.

\paragraph{Input adaptation.} For Moonstone Benchmark, SatMAE, TerraMind, Prithvi, and SpectralGPT+ use input projections adapted to the 30-channel configuration. CROMA receives the four Stokes channels through its synthetic aperture radar branch and the remaining channels through its optical branch. DOFA retains its wavelength-conditioned tokenizer, while Moonstone's radar tokenizer is expanded from two to four channels. NASA--IBM retains its native visual, elevation, and slope inputs and adds three auxiliary modality groups covering the remaining 23 channels. This configuration contains 204.3M encoder parameters, compared with 199.8M for its native adaptation on SomBench. SomBench input projections and spatial token grids follow the task-specific configurations described in the main paper.

\subsection{Pretraining Reconstruction}
\label{sec:supp_reconstruction}

\begin{figure*}[!tp]
    \centering
    \includegraphics[width=\textwidth]{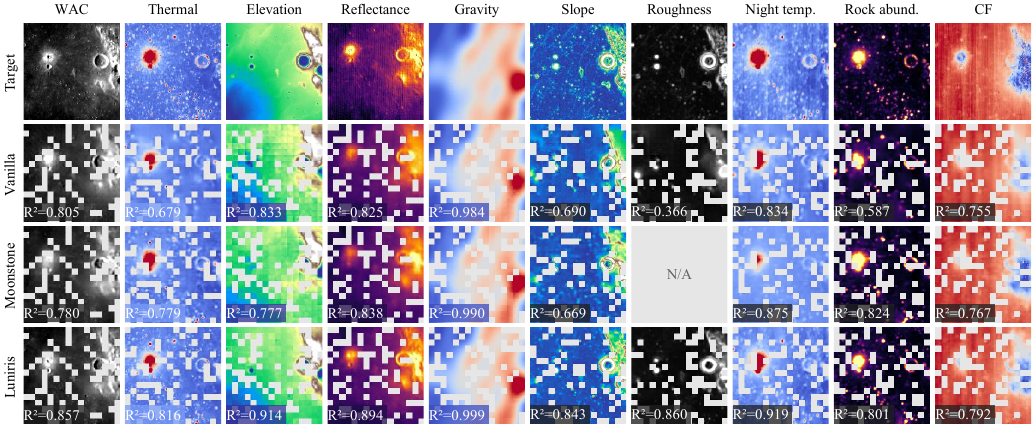}
    \caption{Reconstruction in Region 1 (1/3). Columns show the indicated observation channels; rows show targets, Vanilla MAE, Moonstone MG-MAE, and Luniris. Insets report $R^2$; N/A denotes unavailable reconstructions.}
    \label{fig:recon_supp_r1_p1}
\end{figure*}

\begin{figure*}[!tp]
    \centering
    \includegraphics[width=\textwidth]{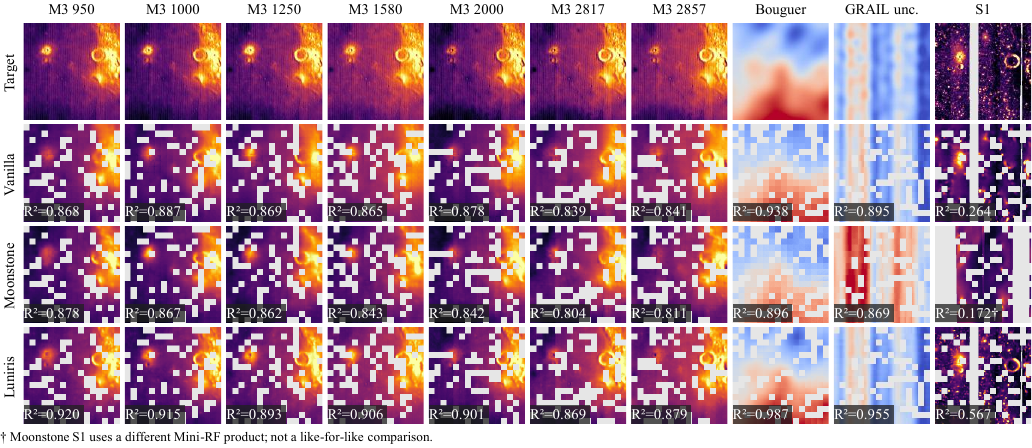}
    \caption{Reconstruction in Region 1 (2/3): additional M$^3$ bands, GRAIL products, and Stokes $S_1$. Layout follows Figure~\ref{fig:recon_supp_r1_p1}. Moonstone's $S_1$ uses a different Mini-RF product.}
    \label{fig:recon_supp_r1_p2}
\end{figure*}

\begin{figure*}[!tp]
    \centering
    \includegraphics[width=\textwidth]{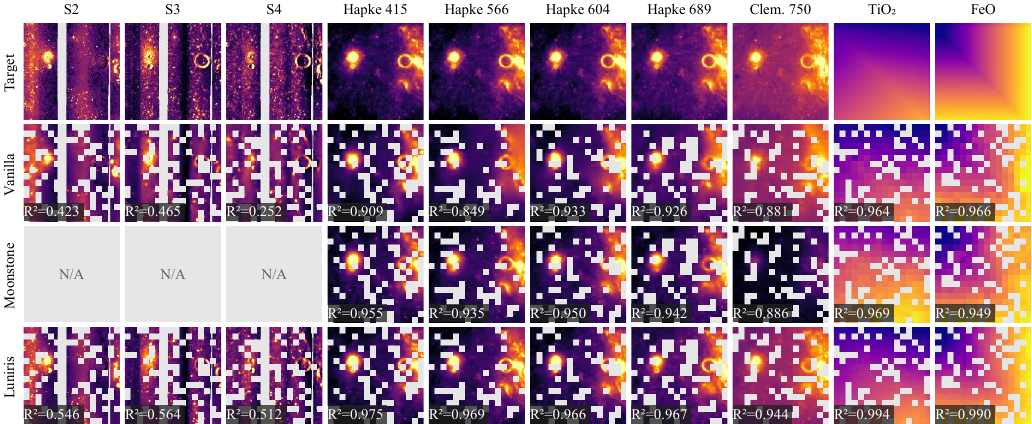}
    \caption{Reconstruction in Region 1 (3/3): Stokes $S_2$--$S_4$, WAC Hapke reflectance, Clementine reflectance, and elemental abundances. Layout follows Figure~\ref{fig:recon_supp_r1_p1}.}
    \label{fig:recon_supp_r1_p3}
\end{figure*}

\begin{figure*}[!tp]
    \centering
    \includegraphics[width=\textwidth]{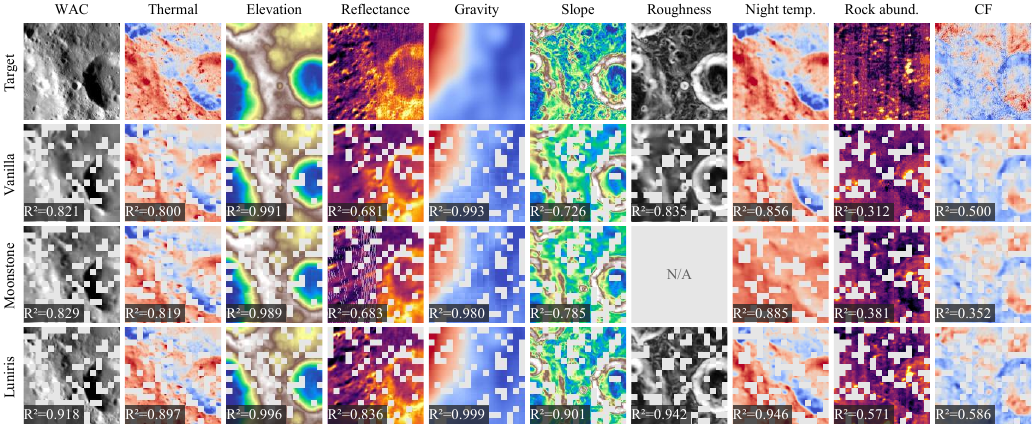}
    \caption{Reconstruction in Region 2 (1/3). Channels, layout, and score conventions follow Figure~\ref{fig:recon_supp_r1_p1}.}
    \label{fig:recon_supp_r2_p1}
\end{figure*}

\begin{figure*}[!tp]
    \centering
    \includegraphics[width=\textwidth]{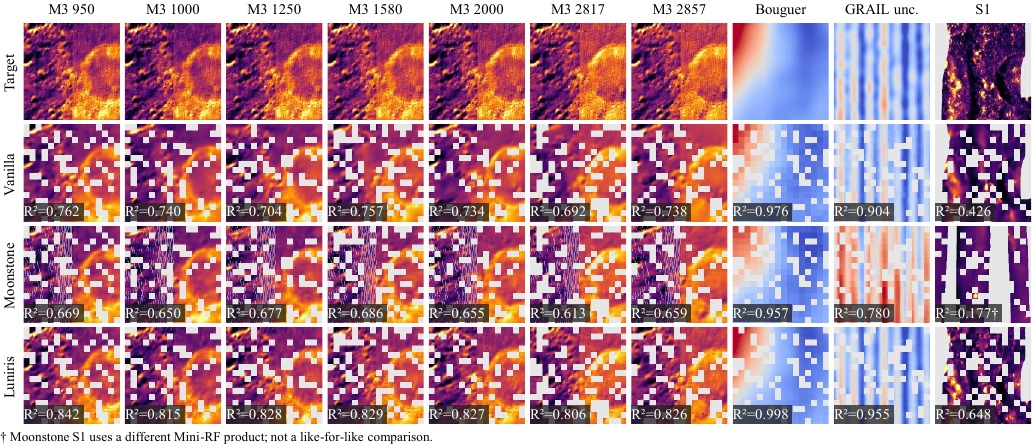}
    \caption{Reconstruction in Region 2 (2/3). Channels and layout follow Figure~\ref{fig:recon_supp_r1_p2}. Moonstone's $S_1$ uses a different Mini-RF product.}
    \label{fig:recon_supp_r2_p2}
\end{figure*}

\begin{figure*}[!tp]
    \centering
    \includegraphics[width=\textwidth]{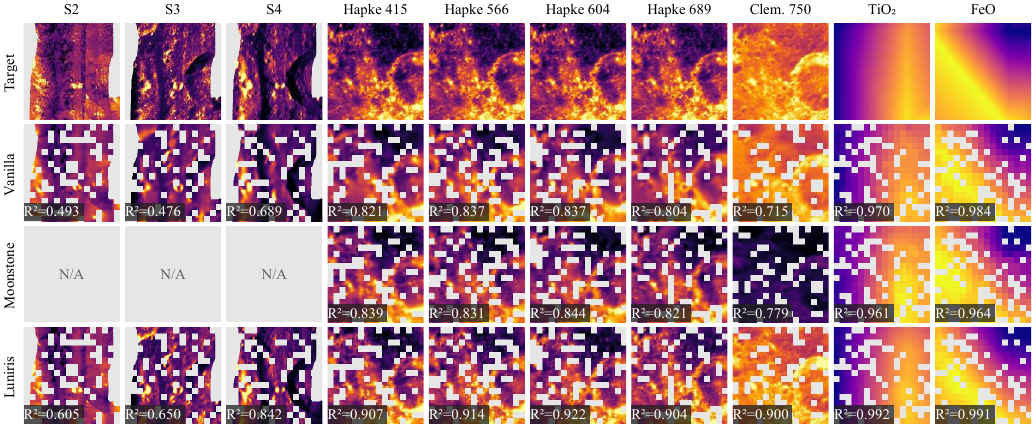}
    \caption{Reconstruction in Region 2 (3/3). Channels, layout, and score conventions follow Figure~\ref{fig:recon_supp_r1_p3}.}
    \label{fig:recon_supp_r2_p3}
\end{figure*}

Figures~\ref{fig:recon_supp_r1_p1}--\ref{fig:recon_supp_r1_p3} and Figures~\ref{fig:recon_supp_r2_p1}--\ref{fig:recon_supp_r2_p3} present all 30 channels in two lunar regions, extending the selected-channel comparison in the main paper. Each figure compares target observations with reconstructions from Vanilla MAE, Moonstone MG-MAE, and Luniris where corresponding outputs are available. Panel scores report the standard coefficient of determination ($R^2$) over valid pixels jointly supported by the reconstruction mask and observation evidence.

Luniris more faithfully recovers localized terrain structures and their surrounding intensity variations. In Region 1, the slope reconstruction preserves the circular high-value rim of the crater near the right edge more distinctly than Vanilla MAE and Moonstone MG-MAE. Its roughness reconstruction also retains the ring-shaped structure against the darker surroundings, whereas Vanilla MAE produces a more diffuse response. The nighttime-temperature channel preserves the compact high-value feature on the left and its contrast with the surrounding field. Region 2 provides a more textured example: Luniris better distinguishes the large crater's dark interior from its bright rim in WAC, while the slope reconstruction follows the curved boundary with clearer local variation. In the thermal and nighttime-temperature channels, it also better preserves the elongated low-value region toward the lower right, rather than blending it into the surrounding higher-value areas.

The optical channels show similar improvements beyond the M$^3$ 750\,nm band displayed in the main paper. Across the additional M$^3$ bands in Region 1, Luniris maintains the compact bright feature on the left and the curved bright structure on the right while preserving their contrast with the darker background. In Region 2, the M$^3$ targets contain more distributed fine-scale intensity variation. Luniris recovers more of this variation around the curved feature and across the lower part of the scene, where Vanilla MAE produces smoother reconstructions and several Moonstone outputs exhibit conspicuous striping. The Hapke and Clementine channels provide complementary examples: Luniris better retains the spatial arrangement of bright structures and the intervening darker areas. These observations extend the optical reconstruction evidence to multiple wavelengths and products.

For slowly varying fields, the advantage appears in the continuity of spatial gradients. Luniris more closely follows the broad transitions in the GRAIL free-air and Bouguer anomaly maps, as well as the TiO$_2$ and FeO abundance maps, with fewer blockwise intensity deviations. It also better preserves the alternating bands in the supplied GRAIL uncertainty maps. Radar reconstruction is compared with Vanilla MAE on the same Stokes targets: Luniris more clearly localizes the strong responses within the valid observation footprint, including the bright structures along the left side of Region 2, while retaining their contrast with adjacent low-response areas. Together, these examples show improved recovery of localized structures, textured observations, and smooth spatial fields across the displayed regions.

\subsection{Results on Moonstone Benchmark}
\label{sec:supp_moonstone}

We evaluate all six Moonstone Benchmark tasks using frozen-encoder probing (LP) and full fine-tuning (FFT). LP freezes the pretrained encoder backbone and trains task-specific prediction heads together with any required input adapters, whereas FFT jointly optimizes the downstream encoder and prediction head. Tables~\ref{tab:supp_moonstone_lp} and~\ref{tab:supp_moonstone_fft} report the complete results under the shared SSP splits. By each task's primary metric, Luniris ranks first on four tasks and among the top two on all six under both settings.

\begin{table*}[!tp]
\centering
\caption{Complete Moonstone Benchmark results under frozen-encoder probing (LP). Bold and underlined values indicate the best and second-best results, ranked before rounding.}
\label{tab:supp_moonstone_lp}
\begingroup
\small
\setlength{\tabcolsep}{1.5pt}
\setlength{\aboverulesep}{0.6pt}
\setlength{\belowrulesep}{0.6pt}
\renewcommand{\arraystretch}{0.95}
\renewcommand{\tabularxcolumn}[1]{m{#1}}

\begin{tabular}{@{}ll*{10}{c}@{}}
\toprule
& & Control & \multicolumn{6}{c}{Earth FMs} & \multicolumn{2}{c}{Lunar FMs} & Ours \\
\cmidrule(l{4pt}r{4pt}){3-3}\cmidrule(l{4pt}r{4pt}){4-9}\cmidrule(l{4pt}r{4pt}){10-11}\cmidrule(l{4pt}r{2pt}){12-12}
Task & Metric & Vanilla~MAE & SatMAE & DOFA & TerraMind & Prithvi & SpectralGPT+ & CROMA & Moonstone & NASA--IBM & \textbf{Luniris} \\
\midrule
\rowcolor{black!5}\multicolumn{12}{l}{\textbf{Moonstone Benchmark --- LP}} \\
\midrule

\multirow{2}{*}{Geo} & \mbox{Acc $\uparrow$} & \underline{0.335} & 0.306 & 0.304 & 0.282 & 0.291 & 0.284 & 0.255 & 0.327 & 0.285 & \textbf{0.379} \\
 & \mbox{mF1 $\uparrow$} & \underline{0.193} & 0.183 & 0.126 & 0.157 & 0.139 & 0.124 & 0.109 & 0.163 & 0.132 & \textbf{0.216} \\
\arrayrulecolor{black!25}\cmidrule[0.25pt]{1-12}\arrayrulecolor{black}

\multirow{2}{*}{Age} & \mbox{Acc $\uparrow$} & 0.533 & 0.484 & 0.526 & 0.506 & 0.501 & 0.489 & 0.491 & \textbf{0.542} & 0.496 & \underline{0.539} \\
 & \mbox{mF1 $\uparrow$} & 0.332 & 0.330 & 0.362 & 0.371 & 0.337 & 0.272 & 0.375 & \underline{0.410} & 0.363 & \textbf{0.428} \\
\arrayrulecolor{black!25}\cmidrule[0.25pt]{1-12}\arrayrulecolor{black}

\multirow{2}{*}{Comp} & \mbox{m$R^2$ $\uparrow$} & \textbf{0.922} & 0.875 & 0.824 & 0.886 & 0.884 & 0.889 & 0.878 & 0.884 & 0.891 & \underline{0.913} \\
 & \mbox{mRMSE $\downarrow$} & \textbf{0.277} & 0.353 & 0.417 & 0.336 & 0.340 & 0.332 & 0.347 & 0.339 & 0.329 & \underline{0.294} \\
\arrayrulecolor{black!25}\cmidrule[0.25pt]{1-12}\arrayrulecolor{black}

\multirow{2}{*}{Therm} & \mbox{m$R^2$ $\uparrow$} & 0.907 & 0.891 & 0.757 & 0.891 & 0.903 & 0.859 & \underline{0.912} & 0.903 & 0.833 & \textbf{0.923} \\
 & \mbox{mRMSE $\downarrow$} & 0.326 & 0.354 & 0.527 & 0.351 & 0.331 & 0.401 & \underline{0.317} & 0.331 & 0.436 & \textbf{0.302} \\
\arrayrulecolor{black!25}\cmidrule[0.25pt]{1-12}\arrayrulecolor{black}

\multirow{2}{*}{Mare} & \mbox{mIoU $\uparrow$} & \underline{0.919} & 0.909 & 0.891 & 0.910 & 0.901 & 0.825 & 0.913 & 0.916 & 0.906 & \textbf{0.920} \\
 & \mbox{mF1 $\uparrow$} & \underline{0.957} & 0.951 & 0.940 & 0.952 & 0.946 & 0.898 & 0.953 & 0.955 & 0.949 & \textbf{0.957} \\
\arrayrulecolor{black!25}\cmidrule[0.25pt]{1-12}\arrayrulecolor{black}

\multirow{2}{*}{Crater} & \mbox{mIoU $\uparrow$} & \underline{0.648} & 0.620 & 0.531 & 0.581 & 0.587 & 0.555 & 0.614 & 0.635 & 0.627 & \textbf{0.668} \\
 & \mbox{mF1 $\uparrow$} & \underline{0.783} & 0.760 & 0.685 & 0.730 & 0.735 & 0.707 & 0.756 & 0.773 & 0.765 & \textbf{0.798} \\
\bottomrule
\end{tabular}

\par\vspace{2pt}
\begin{minipage}{0.93\textwidth}
\small
\textit{Notes.}\ Model variants and metric definitions follow Table~\ref{tab:benchmark_combined}. Hyperparameters and checkpoints are selected using validation-set metrics.
\end{minipage}
\endgroup
\end{table*}

\paragraph{Transfer with a frozen encoder.} The LP results show that Luniris's pretrained representations support both geological discrimination and dense prediction without updating the encoder backbone. On Geo, Luniris achieves an accuracy of 0.379 and macro-F1 of 0.216, compared with 0.327/0.163 for Moonstone MG-MAE and 0.285/0.132 for NASA--IBM. The gains in both metrics support improved geological-unit recognition when classes receive either frequency-based or equal weighting. On Age, Luniris combines the second-highest accuracy with the highest macro-F1 of 0.428, extending this evidence to geological-period classification.

The frozen representations also retain information useful for physical-property estimation and spatial segmentation. On Therm, Luniris achieves the highest macro-$R^2$ of 0.923 and the lowest mRMSE of 0.302, supporting the prediction of thermal properties from non-thermal observations. It also leads on both Mare and Crater in mIoU and macro-F1. These results indicate that the benefit extends from image-level classification to spatially resolved predictions. On Comp, Luniris ranks second in both metrics, behind the same-data Vanilla MAE control and ahead of the other evaluated models.

\begin{table*}[!tp]
\centering
\caption{Complete Moonstone Benchmark results under full fine-tuning (FFT). Bold and underlined values indicate the best and second-best results, ranked before rounding.}
\label{tab:supp_moonstone_fft}
\begingroup
\small
\setlength{\tabcolsep}{1.5pt}
\setlength{\aboverulesep}{0.6pt}
\setlength{\belowrulesep}{0.6pt}
\renewcommand{\arraystretch}{0.95}
\renewcommand{\tabularxcolumn}[1]{m{#1}}

\begin{tabular}{@{}ll*{10}{c}@{}}
\toprule
& & Control & \multicolumn{6}{c}{Earth FMs} & \multicolumn{2}{c}{Lunar FMs} & Ours \\
\cmidrule(l{4pt}r{4pt}){3-3}\cmidrule(l{4pt}r{4pt}){4-9}\cmidrule(l{4pt}r{4pt}){10-11}\cmidrule(l{4pt}r{2pt}){12-12}
Task & Metric & Vanilla~MAE & SatMAE & DOFA & TerraMind & Prithvi & SpectralGPT+ & CROMA & Moonstone & NASA--IBM & \textbf{Luniris} \\
\midrule
\rowcolor{black!5}\multicolumn{12}{l}{\textbf{Moonstone Benchmark --- FFT}} \\
\midrule

\multirow{2}{*}{Geo} & \mbox{Acc $\uparrow$} & \underline{0.362} & 0.309 & 0.336 & 0.299 & 0.281 & 0.295 & 0.282 & 0.356 & 0.342 & \textbf{0.373} \\
 & \mbox{mF1 $\uparrow$} & \underline{0.199} & 0.165 & 0.183 & 0.151 & 0.168 & 0.165 & 0.151 & 0.196 & \textbf{0.208} & 0.190 \\
\arrayrulecolor{black!25}\cmidrule[0.25pt]{1-12}\arrayrulecolor{black}

\multirow{2}{*}{Age} & \mbox{Acc $\uparrow$} & \underline{0.517} & 0.501 & 0.515 & 0.496 & 0.510 & 0.483 & 0.481 & 0.505 & 0.484 & \textbf{0.532} \\
 & \mbox{mF1 $\uparrow$} & 0.368 & 0.362 & 0.355 & 0.321 & \underline{0.398} & 0.328 & 0.381 & 0.391 & \textbf{0.413} & 0.377 \\
\arrayrulecolor{black!25}\cmidrule[0.25pt]{1-12}\arrayrulecolor{black}

\multirow{2}{*}{Comp} & \mbox{m$R^2$ $\uparrow$} & \textbf{0.922} & 0.899 & 0.860 & 0.886 & 0.888 & 0.869 & 0.884 & 0.884 & 0.910 & \underline{0.911} \\
 & \mbox{mRMSE $\downarrow$} & \textbf{0.278} & 0.316 & 0.373 & 0.335 & 0.334 & 0.361 & 0.339 & 0.340 & 0.299 & \underline{0.297} \\
\arrayrulecolor{black!25}\cmidrule[0.25pt]{1-12}\arrayrulecolor{black}

\multirow{2}{*}{Therm} & \mbox{m$R^2$ $\uparrow$} & \underline{0.940} & 0.859 & 0.850 & 0.916 & 0.893 & 0.877 & 0.923 & 0.931 & 0.877 & \textbf{0.941} \\
 & \mbox{mRMSE $\downarrow$} & \textbf{0.262} & 0.402 & 0.415 & 0.309 & 0.348 & 0.376 & 0.295 & 0.282 & 0.374 & \underline{0.267} \\
\arrayrulecolor{black!25}\cmidrule[0.25pt]{1-12}\arrayrulecolor{black}

\multirow{2}{*}{Mare} & \mbox{mIoU $\uparrow$} & 0.919 & 0.924 & 0.921 & 0.918 & 0.918 & 0.828 & 0.911 & \textbf{0.932} & 0.923 & \underline{0.925} \\
 & \mbox{mF1 $\uparrow$} & 0.957 & 0.960 & 0.958 & 0.956 & 0.956 & 0.901 & 0.952 & \textbf{0.964} & 0.959 & \underline{0.960} \\
\arrayrulecolor{black!25}\cmidrule[0.25pt]{1-12}\arrayrulecolor{black}

\multirow{2}{*}{Crater} & \mbox{mIoU $\uparrow$} & 0.656 & 0.614 & 0.572 & 0.603 & 0.610 & 0.604 & 0.617 & 0.631 & \underline{0.666} & \textbf{0.693} \\
 & \mbox{mF1 $\uparrow$} & 0.789 & 0.757 & 0.721 & 0.748 & 0.754 & 0.748 & 0.759 & 0.771 & \underline{0.795} & \textbf{0.816} \\
\bottomrule
\end{tabular}

\par\vspace{2pt}
\begin{minipage}{0.93\textwidth}
\small
\textit{Notes.}\ Model variants and metric definitions follow Table~\ref{tab:benchmark_combined}. Hyperparameters and checkpoints are selected using validation-set metrics.
\end{minipage}
\endgroup
\end{table*}

\paragraph{Adaptation through full fine-tuning.} FFT further improves Luniris's crater segmentation and thermal prediction. Crater mIoU increases from 0.668 under LP to 0.693 under FFT, exceeding NASA--IBM and Moonstone MG-MAE by 2.76 and 6.21 percentage points, respectively. The corresponding macro-F1 of 0.816 also ranks first, showing agreement between the two segmentation metrics. On Therm, macro-$R^2$ increases from 0.923 to 0.941 while mRMSE decreases from 0.302 to 0.267, indicating that task-specific encoder adaptation improves the recovery of the thermal targets. Mare mIoU also increases to 0.925, ranking second. Together with the leading Geo and Age accuracies under FFT, these results show that Luniris provides an effective initialization for supervised adaptation across classification, regression, and segmentation.

\paragraph{Comparison under matched pretraining data.} Luniris outperforms Vanilla MAE, which uses the same pretraining corpus, on five of the six tasks under both LP and FFT by the primary metric. This comparison supports the value of Luniris's pretraining design beyond the choice of lunar training data. Luniris also exceeds every evaluated Earth-pretrained baseline on all six primary metrics in both settings. The complete results therefore extend the main-paper comparison: its advantages appear both when the pretrained backbone is fixed and when it is adapted to individual tasks.

\subsection{Results on SomBench}
\label{sec:supp_sombench}

We evaluate all four SomBench tasks with LoRA and full fine-tuning (FFT). LoRA trains low-rank updates, task heads, and required input adapters while keeping the pretrained backbone weights fixed; FFT jointly optimizes the downstream encoder and task head. Following the input configurations described in the main paper, these experiments assess transfer to datasets not used in pretraining across changes in spatial resolution, modality availability, and patch organization. Tables~\ref{tab:supp_sombench_lora} and~\ref{tab:supp_sombench_fft} provide the complete results. Luniris ranks among the top two on WAC, NAC, and Ice under both adaptation settings.

\begin{table*}[!tp]
\centering
\caption{Complete SomBench results under LoRA adaptation. Bold and underlined values indicate the best and second-best results, ranked before rounding.}
\label{tab:supp_sombench_lora}
\begingroup
\small
\setlength{\tabcolsep}{1.5pt}
\setlength{\aboverulesep}{0.6pt}
\setlength{\belowrulesep}{0.6pt}
\renewcommand{\arraystretch}{0.95}
\renewcommand{\tabularxcolumn}[1]{m{#1}}

\begin{tabular}{@{}ll*{10}{c}@{}}
\toprule
& & Control & \multicolumn{6}{c}{Earth FMs} & \multicolumn{2}{c}{Lunar FMs} & Ours \\
\cmidrule(l{4pt}r{4pt}){3-3}\cmidrule(l{4pt}r{4pt}){4-9}\cmidrule(l{4pt}r{4pt}){10-11}\cmidrule(l{4pt}r{2pt}){12-12}
Task & Metric & Vanilla~MAE & SatMAE & DOFA & TerraMind & Prithvi & SpectralGPT+ & CROMA & Moonstone & NASA--IBM & \textbf{Luniris} \\
\midrule
\rowcolor{black!5}\multicolumn{12}{l}{\textbf{SomBench --- LoRA}} \\
\midrule

\multirow{2}{*}{WAC} & \mbox{mAP $\uparrow$} & 0.168 & 0.170 & 0.174 & 0.172 & 0.175 & 0.156 & 0.171 & 0.177 & \textbf{0.257} & \underline{0.241} \\
 & \mbox{mAP50 $\uparrow$} & 0.325 & 0.340 & 0.351 & 0.336 & 0.336 & 0.309 & 0.335 & 0.351 & \textbf{0.613} & \underline{0.576} \\
\arrayrulecolor{black!25}\cmidrule[0.25pt]{1-12}\arrayrulecolor{black}

\multirow{2}{*}{NAC} & \mbox{mAP $\uparrow$} & 0.103 & 0.099 & 0.118 & 0.092 & 0.098 & 0.077 & 0.108 & 0.083 & \textbf{0.168} & \underline{0.166} \\
 & \mbox{mAP50 $\uparrow$} & 0.264 & 0.242 & 0.285 & 0.221 & 0.245 & 0.192 & 0.269 & 0.203 & \underline{0.496} & \textbf{0.515} \\
\arrayrulecolor{black!25}\cmidrule[0.25pt]{1-12}\arrayrulecolor{black}

\multirow{2}{*}{IMP} & \mbox{mIoU $\uparrow$} & 0.553 & 0.688 & \underline{0.719} & 0.654 & 0.688 & 0.644 & 0.713 & 0.675 & \textbf{0.762} & 0.694 \\
 & \mbox{mF1 $\uparrow$} & 0.633 & 0.789 & \underline{0.815} & 0.755 & 0.788 & 0.744 & 0.811 & 0.776 & \textbf{0.850} & 0.794 \\
\arrayrulecolor{black!25}\cmidrule[0.25pt]{1-12}\arrayrulecolor{black}

\multirow{3}{*}{Ice} & \mbox{RMSE $\downarrow$} & 0.043 & 0.050 & 0.086 & 0.049 & 0.045 & 0.054 & 0.042 & 0.047 & \underline{0.033} & \textbf{0.031} \\
 & \mbox{MAE $\downarrow$} & 0.029 & 0.034 & 0.062 & 0.034 & 0.031 & 0.038 & 0.029 & 0.032 & \underline{0.023} & \textbf{0.021} \\
 & \mbox{$R^2$ $\uparrow$} & 0.975 & 0.967 & 0.900 & 0.968 & 0.973 & 0.961 & 0.976 & 0.970 & \underline{0.985} & \textbf{0.987} \\
\bottomrule
\end{tabular}

\par\vspace{2pt}
\begin{minipage}{0.93\textwidth}
\small
\textit{Notes.}\ Model variants and metric definitions follow Table~\ref{tab:benchmark_combined}. Hyperparameters and checkpoints are selected using validation-set metrics.
\end{minipage}
\endgroup
\end{table*}

\begin{table*}[!tp]
\centering
\caption{Complete SomBench results under full fine-tuning (FFT). Bold and underlined values indicate the best and second-best results, ranked before rounding.}
\label{tab:supp_sombench_fft}
\begingroup
\small
\setlength{\tabcolsep}{1.5pt}
\setlength{\aboverulesep}{0.6pt}
\setlength{\belowrulesep}{0.6pt}
\renewcommand{\arraystretch}{0.95}
\renewcommand{\tabularxcolumn}[1]{m{#1}}

\begin{tabular}{@{}ll*{10}{c}@{}}
\toprule
& & Control & \multicolumn{6}{c}{Earth FMs} & \multicolumn{2}{c}{Lunar FMs} & Ours \\
\cmidrule(l{4pt}r{4pt}){3-3}\cmidrule(l{4pt}r{4pt}){4-9}\cmidrule(l{4pt}r{4pt}){10-11}\cmidrule(l{4pt}r{2pt}){12-12}
Task & Metric & Vanilla~MAE & SatMAE & DOFA & TerraMind & Prithvi & SpectralGPT+ & CROMA & Moonstone & NASA--IBM & \textbf{Luniris} \\
\midrule
\rowcolor{black!5}\multicolumn{12}{l}{\textbf{SomBench --- FFT}} \\
\midrule

\multirow{2}{*}{WAC} & \mbox{mAP $\uparrow$} & 0.153 & 0.157 & 0.166 & 0.154 & 0.151 & 0.144 & 0.146 & 0.163 & \textbf{0.255} & \underline{0.237} \\
 & \mbox{mAP50 $\uparrow$} & 0.309 & 0.304 & 0.324 & 0.302 & 0.302 & 0.290 & 0.301 & 0.323 & \textbf{0.618} & \underline{0.582} \\
\arrayrulecolor{black!25}\cmidrule[0.25pt]{1-12}\arrayrulecolor{black}

\multirow{2}{*}{NAC} & \mbox{mAP $\uparrow$} & 0.081 & 0.085 & 0.101 & 0.079 & 0.090 & 0.074 & 0.085 & 0.093 & \underline{0.164} & \textbf{0.165} \\
 & \mbox{mAP50 $\uparrow$} & 0.206 & 0.209 & 0.248 & 0.199 & 0.219 & 0.189 & 0.216 & 0.228 & \underline{0.500} & \textbf{0.502} \\
\arrayrulecolor{black!25}\cmidrule[0.25pt]{1-12}\arrayrulecolor{black}

\multirow{2}{*}{IMP} & \mbox{mIoU $\uparrow$} & 0.653 & 0.696 & \underline{0.740} & 0.706 & 0.729 & 0.592 & 0.675 & 0.702 & \textbf{0.749} & 0.638 \\
 & \mbox{mF1 $\uparrow$} & 0.754 & 0.794 & \underline{0.833} & 0.803 & 0.823 & 0.686 & 0.775 & 0.800 & \textbf{0.840} & 0.741 \\
\arrayrulecolor{black!25}\cmidrule[0.25pt]{1-12}\arrayrulecolor{black}

\multirow{3}{*}{Ice} & \mbox{RMSE $\downarrow$} & 0.039 & 0.045 & 0.072 & 0.041 & 0.039 & 0.047 & 0.039 & 0.041 & \textbf{0.028} & \underline{0.028} \\
 & \mbox{MAE $\downarrow$} & 0.026 & 0.031 & 0.049 & 0.028 & 0.026 & 0.032 & 0.026 & 0.028 & \textbf{0.018} & \underline{0.019} \\
 & \mbox{$R^2$ $\uparrow$} & 0.980 & 0.972 & 0.929 & 0.977 & 0.980 & 0.970 & 0.980 & 0.977 & \textbf{0.990} & \underline{0.989} \\
\bottomrule
\end{tabular}

\par\vspace{2pt}
\begin{minipage}{0.93\textwidth}
\small
\textit{Notes.}\ Model variants and metric definitions follow Table~\ref{tab:benchmark_combined}. Hyperparameters and checkpoints are selected using validation-set metrics.
\end{minipage}
\endgroup
\end{table*}

\paragraph{Crater detection across observation scales.} Luniris maintains strong detection performance on both multispectral WAC and single-band NAC imagery. On WAC, its mAP reaches 0.241 with LoRA and 0.237 with FFT, exceeding Moonstone MG-MAE's 0.177 and 0.163 and every evaluated Earth-pretrained baseline in the corresponding setting. On NAC, Luniris ranks second in mAP with LoRA and first with FFT, while achieving the highest mAP50 under both settings. These results support the transfer of its pretrained representations to crater localization across different observation scales and spectral inputs. The consistent advantages over Earth-pretrained baselines on both detection tasks further support the value of pretraining tailored to lunar observations.

\paragraph{Transfer to physical-property regression.} Luniris also transfers effectively to ice prospectivity estimation with a different combination of input modalities. Under LoRA, it achieves the best RMSE, MAE, and $R^2$, outperforming both lunar baselines and all evaluated Earth-pretrained models. FFT further reduces its RMSE from 0.031 to 0.028, with all three metrics ranking second to NASA--IBM. Its strong performance under both settings extends the transfer evidence from object localization to dense physical-property prediction.

\paragraph{Segmentation and transfer under matched pretraining data.} The IMP results provide complementary evidence for transfer to single-band segmentation. With LoRA, Luniris achieves an mIoU of 0.694 and macro-F1 of 0.794, compared with 0.553/0.633 for Vanilla MAE and 0.675/0.776 for Moonstone MG-MAE. Thus, adapting a restricted set of parameters is sufficient to obtain useful segmentation representations for irregular mare patches. More broadly, Luniris outperforms the same-data Vanilla MAE control on all four tasks with LoRA and on WAC, NAC, and Ice with FFT. These comparisons support the contribution of its pretraining design to downstream transfer, with advantages that persist across both adaptation settings on detection and regression.

\subsection{Additional Qualitative Comparisons}
\label{sec:supp_qualitative}

\begin{figure*}[!tp]
    \centering
    \includegraphics[width=\textwidth]{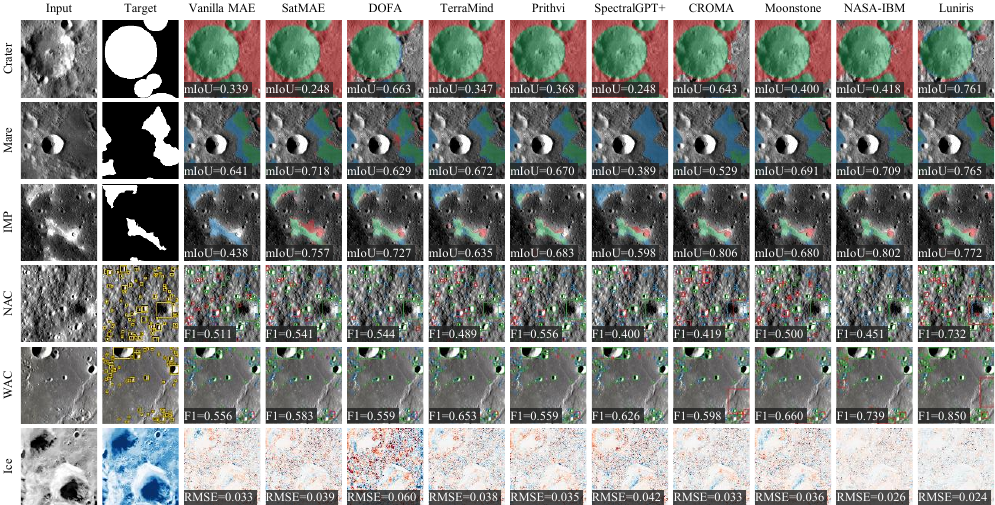}
    \caption{Additional qualitative comparisons, Scene 2 for each task. Rows show Crater, Mare, IMP, NAC, WAC, and Ice. Green/red/blue denote TP/FP/FN; dashed boxes mark missed detections. Ice shows prediction-minus-target residuals, with red/blue indicating overestimation/underestimation. Scores are scene-level; detection F1 uses score and IoU thresholds of 0.5. Model variants follow Table~\ref{tab:benchmark_combined}.}
    \label{fig:supp_downstream_loc02}
\end{figure*}

\begin{figure*}[!tp]
    \centering
    \includegraphics[width=\textwidth]{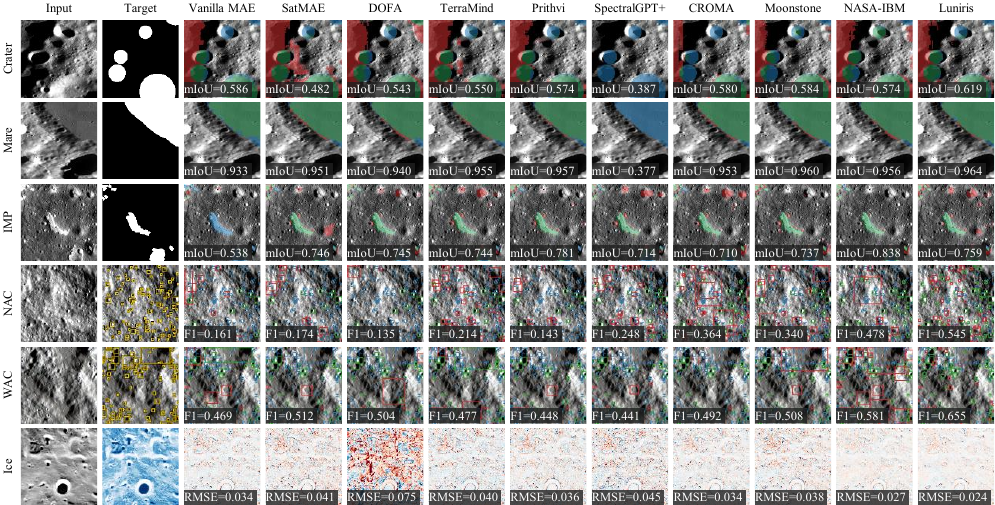}
    \caption{Additional qualitative comparisons, Scene 3 for each task. Task order, color conventions, model variants, and scene-level metrics follow Figure~\ref{fig:supp_downstream_loc02}.}
    \label{fig:supp_downstream_loc03}
\end{figure*}

Figures~\ref{fig:supp_downstream_loc02} and~\ref{fig:supp_downstream_loc03} extend Figure~\ref{fig:downstream} with two additional selected test scenes for each task. Scene indices are assigned within each task and do not indicate shared geographic locations across tasks. All models use the same input, target, and evaluation region within each comparison. These examples illustrate how the quantitative differences appear in foreground delineation, crater localization, and spatial regression errors.

\paragraph{Foreground delineation.} The additional Crater examples highlight improved separation of annotated craters from their surroundings. In Scene 2, a large crater occupies much of the image, with smaller annotated craters near the boundaries. Moonstone MG-MAE and NASA--IBM recover the large crater but also label broad surrounding areas as foreground. Luniris suppresses much of this excess foreground while retaining the main crater, producing a more distinct circular outline. Scene 3 contains strong illumination contrasts and several dark crater interiors along the left side. Luniris recovers more of the two annotated craters at middle-left than the lunar baselines, reducing the missed foreground within these objects.

The Mare examples show complementary improvements in region continuity and boundary placement. In Scene 2, Luniris more fully recovers the narrow connection between the upper and lower portions of the right-hand mare region and reduces omissions along the left image boundary. In Scene 3, it closely follows the diagonal boundary separating the annotated mare region from the surrounding textured terrain, with less false-positive spillover along the central boundary than Moonstone MG-MAE and NASA--IBM. The IMP examples further illustrate recovery of irregular shapes: Luniris captures more of the elongated patch's lower-right tip in Scene 2 and recovers much of the bottom-right patch missed by Moonstone MG-MAE in Scene 3. These localized improvements complement the aggregate segmentation results.

\paragraph{Crater localization.} On NAC, the main advantage is more complete detection of annotated craters in densely textured terrain. In Scene 2, Luniris recovers numerous small craters across the central and lower portions of the image that remain blue missed-detection boxes in the lunar baselines. It matches 60 of the 79 annotated craters, compared with 37 for Moonstone MG-MAE and 23 for NASA--IBM at the displayed operating point. Scene 3 shows a similar recovery of small targets, particularly in the lower half, extending this observation to another terrain configuration.

The WAC examples show improvements in both target recovery and box localization. In Scene 2, Luniris detects more of the small annotated craters along the left side and near the lower-right cluster while retaining the prominent large craters near the upper edge. In Scene 3, Moonstone MG-MAE and NASA--IBM produce several oversized false-positive boxes spanning broad terrain regions. Luniris avoids many of these boxes and recovers more individual annotated craters. Its WAC predictions contain both more true positives and fewer false positives than NASA--IBM in the two displayed scenes, supporting improved localization rather than simply a larger number of detections.

\paragraph{Spatial regression errors.} The Ice examples show reduced residual structure across two additional target fields. In Scene 2, Moonstone MG-MAE produces pronounced alternating positive and negative residuals across the textured upper half and around the large lower crater. Luniris reduces these patterns, leaving a larger proportion of the scene close to zero error. In Scene 3, it similarly attenuates the residual texture across the upper-left terrain and the central region. NASA--IBM also produces relatively low-error maps, while Luniris achieves lower scene-level RMSE in both examples: 0.024 versus 0.026 in Scene 2 and 0.024 versus 0.027 in Scene 3. Together, the spatial maps and corresponding errors support more accurate recovery of the target field across these selected scenes.

\end{document}